\documentclass{article}

\usepackage[preprint]{main}

\usepackage[utf8]{inputenc} % allow utf-8 input
\usepackage[T1]{fontenc}    % use 8-bit T1 fonts
\usepackage{hyperref}       % hyperlinks
\usepackage{url}            % simple URL typesetting
\usepackage{booktabs}       % professional-quality tables
\usepackage{amsfonts}       % blackboard math symbols
\usepackage{nicefrac}       % compact symbols for 1/2, etc.
\usepackage{microtype}      % microtypography
\usepackage[table]{xcolor}  % colors (table shading)
\usepackage{amsmath}
\usepackage{graphicx}
\usepackage{wrapfig}
\usepackage{multirow}
\usepackage{tikz}
\usepackage{pgfplots}
\pgfplotsset{compat=1.18}
\usetikzlibrary{shapes.geometric, arrows.meta, positioning, fit, calc, backgrounds}
\usepackage[utf8]{inputenc} % 或者是 xeCJK 处理中文
\usepackage{xparse}
\usepackage{tcolorbox}
\tcbuselibrary{breakable,skins}

\usepackage{pifont}

\definecolor{promptgray}{gray}{0.95}
\definecolor{lightgreen}{HTML}{c9dfb7}

\makeatletter
\@ifundefined{nolinenumbers}{%
  \newenvironment{nolinenumbers}{}{}%
}{}
\makeatother

\NewDocumentEnvironment{promptbox}{O{}}{%
  \par
  \begin{nolinenumbers}
  \begin{tcolorbox}[
    enhanced jigsaw,
    breakable,
    colback=promptgray,
    colframe=black,
    coltext=black,
    boxrule=0.5pt,
    sharp corners,
    left=6pt,
    right=6pt,
    top=6pt,
    bottom=6pt,
    fonttitle=\bfseries,
    title={#1},
    before skip=1em,
    after skip=1em
  ]
}{%
  \end{tcolorbox}
  \end{nolinenumbers}
  \par
}

\newcommand{\pvar}[1]{\texttt{\{\{ #1 \}\}}}

\title{Deep Research as Rubric for Reinforcement Learning}

\author{%
  Wangyi Mei\textsuperscript{1,2},
  Zhouhong Gu\textsuperscript{2},
  Zhenhan Bai\textsuperscript{2,3},
  Yin Cai\textsuperscript{1,2},
  Lefan Zhang\textsuperscript{2},
  Zhenxin Ding\textsuperscript{2},\\
  \bfseries Bo Chen\textsuperscript{2},
  Yan Gao\textsuperscript{2},
  Yi Wu\textsuperscript{2},
  Yao Hu\textsuperscript{2},
  Jiaqing Liang\textsuperscript{1},
  Deqing Yang\textsuperscript{1,*} \\[6pt]
  \textsuperscript{1}Fudan University,
  \textsuperscript{2}Xiaohongshu Inc.,
  \textsuperscript{3}Beijing University of Posts and Telecommunications \\[4pt]
  \texttt{\{wymei24@m.,ycai25@m.,yangdeqing@,liangjiaqing@\}fudan.edu.cn}, \\
  \texttt{zhenhanbai2023@bupt.edu.cn} \\
  \texttt{\{guzhouhong1,zhanglefan,dingzhenxin,chenbo\}@xiaohongshu.com}, \\
  \texttt{\{wanjianyi,luyun2,xiahou\}@xiaohongshu.com} \\[4pt]
  Code: \textcolor{blue}{\url{https://github.com/meiotoufa/DR-Rubric}}
}

\begin{document}

\maketitle

\begin{abstract}
Open-ended reasoning and long-form generation tasks lack reliable automatic verification signals for reward-based policy optimization. Rubrics offer a promising alternative, but existing approaches treat them as given artifacts---either hand-crafted or prompt-generated---and often miss the task-specific, knowledge-intensive dimensions that matter most, distorting the reward signal. Our key observation is that \emph{rubric construction is itself a research problem}: identifying what makes a response correct or insightful requires discovering and synthesizing external knowledge. We propose \textbf{Deep Research as Rubric (DR-rubric)}, a two-stage framework for constructing such rubrics. Stage~I elicits domain facts, structural constraints, and failure modes through iterative multi-turn agentic search; Stage~II distills this evidence into atomic, independently verifiable constraints for GRPO-based policy optimization. Because the model under training can serve as its own rubric generator, DR-rubric-8B supports \emph{bootstrap rubric generation} without frontier-model assistance. We evaluate on 6 benchmarks spanning agentic research and expert reasoning. Experiments show that DR-Rubric achieves strong competitive performance with only 1K–3K training instances, where GPT-5-generated rubrics particularly benefit breadth coverage on agentic tasks, Gemini-generated rubrics yield the most balanced performance across agentic and expert reasoning tasks, and bootstrap rubrics exhibit a specialization-to-rebalancing evolution achieving the best overall performance at the third iteration. Results demonstrate that reframing rubric construction from static evaluation templates into an evidence-driven research process yields more scalable, fine-grained reward signals for open-ended tasks.
\end{abstract}
\section{Introduction}
\label{sec:intro}

Large language models (LLMs) have demonstrated remarkable capabilities in complex tasks such as open-ended reasoning and long-form generation. To truly unlock and harness these capabilities, reinforcement learning has become an indispensable component of the post-training stage. However, unlike tasks with clear ground truth such as mathematical proofs, open-ended problems lack definitive reference answers, making it difficult to produce automatic verification signals for policy optimization. \emph{Rubrics}---structured evaluation criteria that decompose holistic quality judgments into fine-grained, interpretable dimensions---offer an effective pathway to address this absence of explicit rewards~\citep{rubrics_as_rewards, chasing_the_tail, breaking_exploration_bottleneck}. By rewarding each satisfied constraint separately, the feedback provides partial credit and gives reinforcement learning a dense and stable training signal.

Despite the strong potential of rubric-based approaches, existing construction methods typically treat rubrics as static, given artifacts, relying heavily on expensive manual authoring or simple prompt-based generation from large models~\citep{healthbench}. While expert-authored rubrics carry high authority, this labor-intensive approach clearly lacks long-term scalability and maintainability in the face of massive, highly diverse training data distributions. Turning to prompt-based generation instead confines rubric quality to the model's internal parametric priors, yielding rubrics that typically assess only surface-level textual coherence and clarity~\citep{openrubrics, auto_rubric}. This generic generation approach, lacking grounding in external factual evidence, critically misses the knowledge-intensive, task-specific constraint dimensions that truly determine response quality. How to automatically generate such ``one-to-one'' customized evaluation criteria at scale constitutes the core technical challenge in this area.

\begin{figure}[t]
\centering
\includegraphics[width=\textwidth]{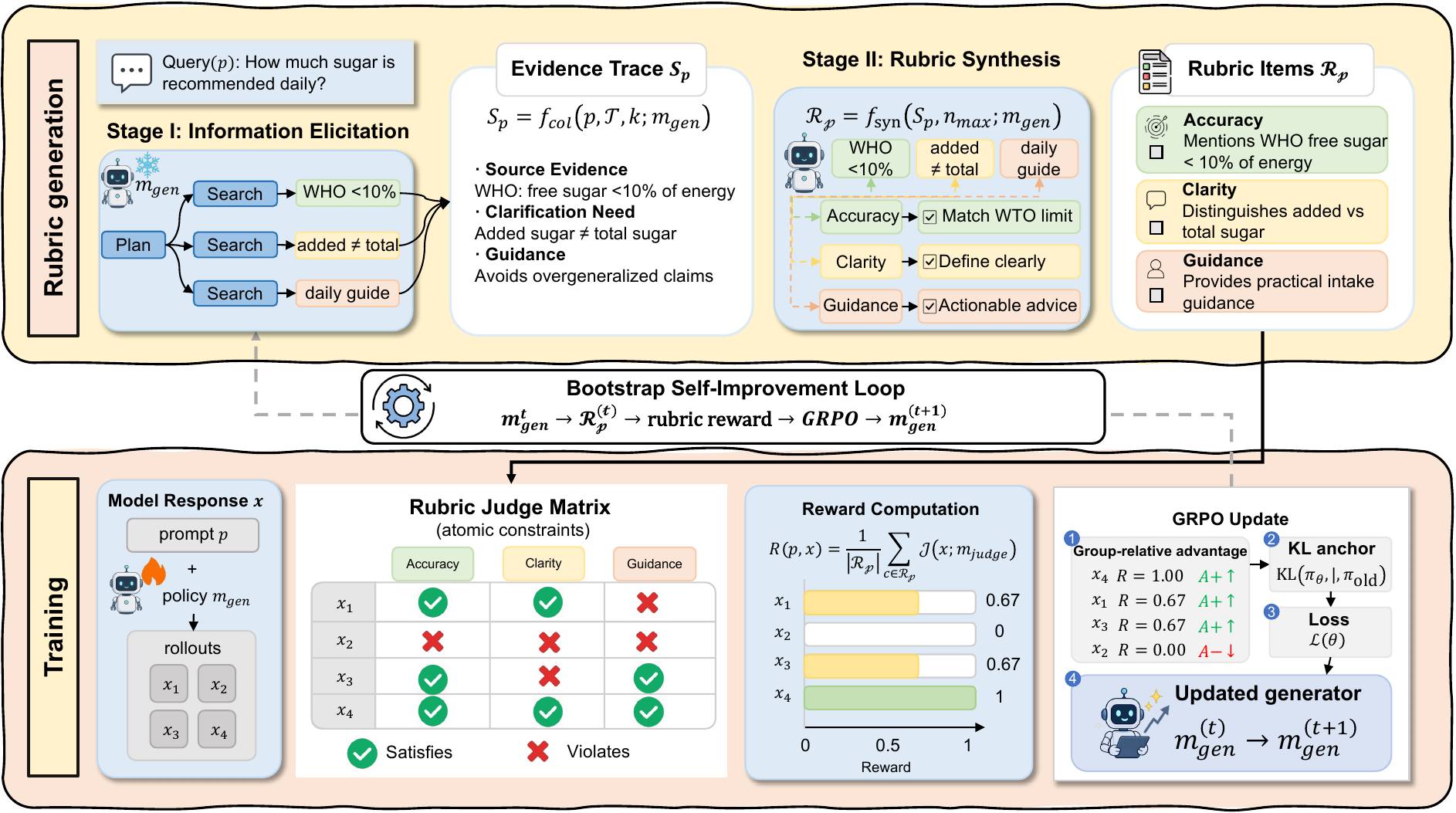}
\caption{Overview of DR-Rubric. Given a task prompt $p$, evidence collection $f_{col}$ conducts agentic research to build a structured evidence trace $\mathcal{S}_p$; constraint synthesis $f_{syn}$ extracts atomic constraints to form the rubric $\mathcal{R}_p$. Rewards are computed per constraint and aggregated for GRPO training. The dashed arrow indicates \emph{bootstrap rubric generation}: the trained policy $\pi_{\theta_{t+1}}$ replaces the external research model in subsequent iterations.}
\label{fig:dr_rubric_overview}
\end{figure}

Our key observation is that \emph{rubric construction is itself a research problem}. Identifying what makes a response correct, comprehensive, or insightful for a given task requires discovering, gathering, and synthesizing relevant knowledge from external sources---precisely the kind of structured inquiry that characterizes deep research. Based on this insight, we propose \textbf{Deep Research as Rubric (DR-Rubric)}, a systematic framework that reframes rubric construction as a two-stage, evidence-driven process. Deep research is an autonomous knowledge acquisition mechanism that conducts iterative multi-turn retrieval and dynamic synthesis of complex external information through agentic interaction. Incorporating this mechanism into the rubric generation process enables the model to actively collect, for each specific query, the objective facts, structural constraints, and common failure patterns needed to formulate evaluation criteria. In \emph{Stage~I: Information Elicitation}, an agentic research model conducts iterative multi-turn search to build a comprehensive evidence trace. In \emph{Stage~II: Rubric Synthesis}, the accumulated evidence is transformed into a formal rubric consisting of atomic, programmatically verifiable constraints that can be evaluated independently against candidate responses. Figure~\ref{fig:dr_rubric_overview} illustrates the full pipeline.

To validate the effectiveness of DR-Rubric, we conduct extensive empirical analysis across 6 benchmark datasets spanning diverse research and reasoning tasks. Among 8B-scale models, DR-Rubric-8B with GPT-5-generated rubrics achieves strong agentic task performance, outperforming all 8B baselines on ResearchQA and DeepResearchBench, while Gemini-generated rubrics yield the most balanced profile, leading on LocalSearchBench and GPQA. More importantly, we explore and confirm the model's capacity for \emph{bootstrap rubric generation}---the ability to self-generate customized rubrics. We find that the first bootstrap step produces strong reasoning performance (GPQA 57.0, MMLU 84.0), surpassing both benchmark-native and GPT-5-assisted rubrics on expert reasoning benchmarks, as self-generated rubrics closely match the model's capability boundary and induce stronger optimization pressure. However, this initial step also triggers capability specialization toward reasoning at the expense of agentic tasks. Subsequent bootstrap iterations effectively repair this imbalance: Bootstrap Step~3 surpasses GPT-5-assisted quality across expert reasoning benchmarks (MMLU-Pro 78.0, MMLU 85.3) without any external model.

Our main contributions are as follows:
\textbf{First}, we show why open-ended task evaluation requires customized rubrics and how to frame rubric construction as an evidence-grounded deep research process, linking knowledge synthesis to the design of evaluation criteria.
\textbf{Second}, we propose DR-Rubric, a two-stage framework comprising Information Elicitation and Rubric Synthesis, that transforms task-specific external evidence into programmatically verifiable, fine-grained constraints.
\textbf{Third}, we demonstrate the feasibility of bootstrap rubric generation and reveal a ``specialization-to-rebalancing'' dynamic across bootstrap iterations, where initial self-generated rubrics drive strong reasoning gains through capability specialization, and subsequent iterations progressively recover agentic task performance through structural rebalancing.

\section{Related Work}

\paragraph{Rubric-based rewards for open-ended post-training.}
For tasks without automatic verifiers---open-ended reasoning, long-form synthesis, medical question answering---recent work converts rubrics into structured reward signals that guide post-training. Decomposing judgments into interpretable criteria has been shown to yield more fine-grained and stable training signals than scalar or black-box preference rewards~\citep{rubrics_as_rewards}, and this idea has been extended to long-tail evaluation in LLM alignment~\citep{chasing_the_tail}. Rubric dimensions can also serve as intermediate objectives in reinforcement learning, mitigating reward sparsity and guiding exploration in complex reasoning settings~\citep{breaking_exploration_bottleneck, DBLP:journals/corr/abs-2508-12790}. Large-scale expert-authored rubrics for medical QA further illustrate both the power and the prohibitive cost of manual rubric construction~\citep{healthbench}. A common thread is that these methods \emph{assume the rubric itself is already available or can be directly obtained}: the research problem is how to use rubrics as reward, not how to construct them. Our work shifts attention from \emph{rubric usage} to \emph{rubric construction}, treating the design of task-specific evaluation criteria as a first-class learning problem.

\paragraph{Automated rubric generation with external evidence.}
To reduce reliance on manual specification, a growing body of work explores automated rubric design. One direction uses LLMs to propose constraint-enhanced rewards that enforce format, coherence, or rule compliance~\citep{ace_rl}. Another synthesizes rubric criteria at scale via LLM generation and filtering~\citep{openrubrics}, dynamically evolves rubric dimensions during training~\citep{dr_tulu_rler}, or derives evaluation dimensions from pairwise comparison data~\citep{online_rubric_elicitation}. These approaches improve scalability over expert-authored rubrics, but they remain largely confined to the model's parametric priors: the generated criteria may be structurally valid yet weakly grounded in the specific factual and failure-mode structure of each task. As a result, automatically generated rubrics tend to emphasize generic surface-level qualities (e.g., clarity, coherence, relevance) while under-specifying knowledge-intensive or task-critical dimensions. DR-Rubric addresses this by decomposing rubric construction into \emph{information elicitation} and \emph{rubric synthesis}, so that criteria are derived from an explicit evidence trace gathered via agentic research rather than from prompt-only generation.

\paragraph{Self-generated feedback and bootstrapped reward learning.}
Self-improvement methods demonstrate that models can provide feedback or supervision for their own training. Early work introduced self-critiques derived from model-generated principles to iteratively refine responses~\citep{bai2022constitutional}. More recent approaches further show that a model can act as both the generator and the evaluator, with self-generated feedback substituting for external supervision~\citep{chen2024self, yuan2024self}. However, the feedback in these approaches is typically scalar, preference-like, or principle-based---it tells the model \emph{whether} a response is good but does not specify \emph{what task-specific standards} it should satisfy. In contrast, DR-Rubric studies self-generated rubrics as structured, constraint-level reward specifications. The key question we pursue is not only whether a model can judge itself, but whether it can construct increasingly useful task-specific evaluation criteria for its own post-training---a capacity we operationalize through bootstrap rubric generation, where the trained policy also serves as the rubric-generating research agent in a self-improving loop.

\section{The DR-Rubric Framework}
\label{sec:method}

DR-Rubric transforms rubric construction from reliance on model-internal priors to a rigorous derivation grounded in external evidence. The framework comprises three sequential components: automated rubric generation (\S\ref{subsec:rubric_generation}), reinforcement learning with rubrics (\S\ref{subsec:rl_with_rubrics}), and bootstrap rubric generation (\S\ref{subsec:bootstrap}). Table~\ref{tab:notation} defines all core symbols used throughout.

\subsection{Rubric Generation}
\label{subsec:rubric_generation}

The rubric generation stage constructs a tailored rubric $\mathcal{R}_p$ for each query $p$, backed by external factual evidence. It consists of two steps: evidence collection and constraint synthesis.

\paragraph{Evidence Collection.}
Evidence collection takes as input a query $p$, an external tool set $\mathcal{T}$, and a maximum interaction budget $k$, and is driven by a generation model $m_{gen}$ that conducts multi-round retrieval and verification. The output is an evidence trace $\mathcal{S}_p$ containing domain facts, structural constraints, and common failure modes:
\begin{equation}
    \mathcal{S}_p = f_{col}(p, \mathcal{T}, k;\ m_{gen}).
\end{equation}
The budget $k$ caps the number of agent--environment interaction rounds, controlling the computational cost of the collection process.

Rather than assuming that evaluation criteria can be inferred directly from $p$, DR-Rubric treats missing task knowledge as a latent variable that must be explicitly acquired. The research process is iterative: at each step, the agent performs query formulation, evidence acquisition, and query refinement, where later queries are conditioned on earlier findings. This promotes coverage and reduces omission of critical constraints.

This formulation differs from standard retrieval-augmented generation in two key aspects. First, the objective is \emph{requirement discovery} rather than answer production: exploration targets information that constrains evaluation rather than content that directly solves the task. Second, the iterative structure allows later queries to build on earlier findings, yielding an evidence trace $\mathcal{S}_p$ that serves as an explicit intermediate representation bridging task prompts and evaluation criteria.

\paragraph{Constraint Synthesis.}
An atomic constraint $c$ is an indivisible single evaluation criterion. The rubric $\mathcal{R}_p$ is defined as the set of atomic constraints tailored to query $p$:
\begin{equation}
    \mathcal{R}_p = \{c_1, c_2, \dots, c_n\}.
\end{equation}
Constraint synthesis distills the unstructured evidence trace $\mathcal{S}_p$ into this form, with $n_{max}$ capping the rubric size to prevent redundancy and noise from overly large constraint sets:
\begin{equation}
    \mathcal{R}_p = f_{syn}(\mathcal{S}_p, n_{max};\ m_{gen}).
\end{equation}
Each constraint is phrased in natural language as either an affirmative requirement (``the response should cover at least two efficient attention methods'') or a negation requirement (``the response should not claim that standard attention is $O(n \log n)$''). Both types reside in a single constraint set with no categorical distinction or differential weighting---they differ only in linguistic form.

Crucially, constraints are not heuristically authored but are traceable to specific elements in $\mathcal{S}_p$. Together, evidence collection and constraint synthesis ensure that evaluation criteria derive from explicit external facts rather than parametric biases accumulated during pretraining.

\begin{table}[t]
\caption{Core notation for DR-Rubric.}
\centering
\small
\setlength{\tabcolsep}{4pt}
\renewcommand{\arraystretch}{1.1}
\begin{tabular}{c|l|l}
\hline
\textbf{Symbol} & \textbf{Definition} & \textbf{Description} \\
\hline
$p$                   & Task Prompt           & A raw query instance from the training set \\
$m_{gen}$             & Generation Model  & The model driving deep research and rubric synthesis \\
$f_{col}$             & Evidence Collection   & Multi-round deep-research evidence gathering process \\
$\mathcal{S}_p$       & Evidence Trace        & Set of facts and structural patterns collected for $p$ \\
$f_{syn}$             & Constraint Synthesis  & Process converting evidence traces into normalized constraints \\
$\mathcal{R}_p$       & Rubric                & Full set of atomic constraints tailored to $p$ \\
$c$                   & Atomic Constraint     & An indivisible single evaluation criterion in the rubric \\
$x$                   & Response              & Text generated by the model given $p$ \\
$m_{judge}$           & Judge Model           & Model performing atomic constraint verification \\
$J_c(x; m_{judge})$   & Judge Predicate       & Discrete indicator of whether $x$ satisfies constraint $c$ \\
$R(p, x)$             & Reward                & Arithmetic mean of scores across all constraints \\
$\pi_\theta$          & Policy Model          & The main model undergoing parameter optimization \\
$\pi_{\theta_{old}}$  & Old Policy            & Policy snapshot before the current update step \\
$\pi_{ref}$           & Reference Model       & Initial base model constraining update magnitude \\
$G$                   & Group Size            & Number of responses sampled per prompt in GRPO \\
$k$                   & Max Steps             & Upper limit on agent--environment interactions during research \\
$n_{max}$             & Max Constraints       & Maximum number of atomic constraints per rubric \\
$\epsilon$            & Clip Coefficient      & Probability ratio clipping range in the PPO objective \\
$\beta$               & KL Penalty            & Regularization weight controlling policy divergence \\
\hline
\end{tabular}
\label{tab:notation}
\end{table}

\subsection{Reinforcement Learning with Rubrics}
\label{subsec:rl_with_rubrics}

The tailored rubric $\mathcal{R}_p$ is converted into a dense reward signal to drive parameter optimization of the policy model $\pi_\theta$. This involves reward computation and policy update.

\paragraph{Reward Computation.}
The judge model $m_{judge}$ independently scores each atomic constraint $c \in \mathcal{R}_p$ with a discrete judgment $J_c(x; m_{judge}) \in \{0, 1\}$, determining whether response $x$ satisfies the constraint. The final reward $R(p, x)$ is the arithmetic mean over all constraint scores, aggregating multi-dimensional quality assessment into a single reward signal:
\begin{equation}
R(p, x) = \frac{1}{|\mathcal{R}_p|} \sum_{c \in \mathcal{R}_p} J_c(x;\ m_{judge}).
\end{equation}
This design distributes supervision across atomic evaluation criteria rather than assigning a single holistic score. Responses receive partial credit through selective constraint satisfaction, producing denser and more informative feedback than task-level binary rewards. Because each component corresponds to an interpretable requirement derived from $\mathcal{S}_p$, the reward remains grounded in evidence-backed task knowledge.

\paragraph{Policy Update.}
Training uses GRPO to update the policy model $\pi_\theta$. For each query $p$, $G$ responses $\{x_1, \dots, x_G\}$ are sampled from the current policy, and the group-normalized reward serves as the advantage estimate $A_i$, eliminating reward-scale differences across queries:
\begin{equation}
A_i = \frac{R(p, x_i) - \text{mean}(\{R(p, x_j)\}_{j=1}^G)}{\text{std}(\{R(p, x_j)\}_{j=1}^G)}.
\end{equation}
A token-level KL divergence approximation prevents the policy $\pi_\theta$ from drifting too far from the reference model $\pi_{ref}$:
\begin{equation}
\text{KL}_{prac}(x_i) = \frac{1}{|x_i|} \sum_{j=1}^{|x_i|} \left( \log \pi_\theta(x_{i,j} \mid p, x_{i,<j}) - \log \pi_{ref}(x_{i,j} \mid p, x_{i,<j}) \right).
\end{equation}
The final GRPO objective combines the clipped probability-ratio objective with the KL penalty, where $\epsilon$ controls the clipping range and $\beta$ is the KL penalty coefficient:
\begin{equation}
\mathcal{L}(\theta) = \frac{1}{G} \sum_{i=1}^{G} \left( \min \left( \frac{\pi_\theta(x_i \mid p)}{\pi_{\theta_{old}}(x_i \mid p)} A_i,\ \text{clip}\!\left(\frac{\pi_\theta(x_i \mid p)}{\pi_{\theta_{old}}(x_i \mid p)},\ 1{-}\epsilon,\ 1{+}\epsilon\right) A_i \right) - \beta \cdot \text{KL}_{prac}(x_i) \right).
\end{equation}

\subsection{Bootstrap Rubric Generation}
\label{subsec:bootstrap}

The automated rubric generation stage relies on an external high-performance model $m_{gen}$ to conduct deep research. In practice, however, continuously calling such closed-source models incurs significant cost and limits scalability. Bootstrap generation breaks this dependency: as the policy model $\pi_\theta$ improves its reasoning capabilities, it gradually becomes capable of assuming the rubric generation role, enabling the entire training pipeline to evolve without external model supervision.

In bootstrap iteration $t$, the current policy $\pi_{\theta_t}$ directly takes over the evidence collection and constraint synthesis responsibilities originally assigned to $m_{gen}$, generating tailored rubrics for new queries:
\begin{equation}
\mathcal{R}_{p,t} = f_{syn}\!\left(f_{col}(p, \mathcal{T}, k;\ \pi_{\theta_t}),\ n_{max};\ \pi_{\theta_t}\right).
\end{equation}
The resulting rubrics then serve as the reward source driving the next round of policy update:
\begin{equation}
\pi_{\theta_{t+1}} = \text{GRPO}(\pi_{\theta_t},\ \mathcal{R}_{p,t}).
\end{equation}
This recurrence constitutes a self-improvement loop bounded by the model's own reasoning capacity: as $\pi_{\theta_t}$ improves, its generated rubrics also improve in quality, providing more precise supervision signals for the next training round. We evaluate this loop empirically in Section~\ref{sec:experiments}, comparing single-step and multi-step bootstrap variants against the GPT-5- and Gemini-assisted baselines.

\section{Experiments}
\label{sec:experiments}

\subsection{Setup}

We evaluate on six benchmarks spanning agentic tasks (ResearchQA~\citep{researchqa}, DeepResearchBench~\citep{deepresearchbench}, LocalSearchBench~\citep{localsearchbench}) and expert reasoning (GPQA~\citep{gpqa}, MMLU-Pro~\citep{mmlu_pro}, MMLU~\citep{MMLU}). All evaluations use a shared mock retrieval backend for reproducibility; Appendix~\ref{appendix:live_api_validation} verifies consistency with live web search. Qwen3-8B-SFT is obtained by fine-tuning Qwen3-8B on 1K ReAct-format trajectories, each sampled once from Qwen3-32B-FP8 using the same task prompts as RL training and following an identical rollout protocol. All DR-Rubric-8B models are initialized from this Qwen3-8B-SFT model. This SFT stage is used to align the model with the multi-turn tool-interaction format and serves as the shared Stage-0 initialization; the subsequent DR-Rubric stage optimizes the model with GRPO on 1K instances per step using rubric-derived rewards. For bootstrap variants, each step trains on the same 1K-instance set from the previous step's checkpoint, yielding 2K and 3K cumulative RL instances for BS-2 and BS-3 respectively. Unless otherwise specified, we use group size $G{=}4$, clip range $\epsilon{=}0.28$, KL coefficient $\beta{=}0.001$, learning rate $5\times10^{-6}$, and 5 epochs on 16$\times$H800 GPUs. Full benchmark, data, and implementation details are in Appendices~\ref{sec:benchmark_descriptions},~\ref{appendix:live_api_validation}, and~\ref{appendix:sample_size}.

\paragraph{Baselines.}
We compare against Qwen2.5-7B, Qwen3-8B~\citep{DBLP:journals/corr/abs-2505-09388}, DR-Tulu-SFT-8B and DR-Tulu-RL-8B~\citep{dr_tulu_rler}, Search-R1-7B~\citep{search_r1}, and WebExplorer-8B~\citep{webexplorer}. Our variants differ only in rubric source: benchmark-native, GPT-5-generated~\citep{DBLP:journals/corr/abs-2601-03267}, Gemini-generated~\citep{googledeepmind2026gemini31pro}, or bootstrap-generated rubrics from Steps~1--3.

\subsection{Main Results}
\label{subsec:main_results}
\newcommand{\up}[1]{\textsuperscript{\textcolor{green!50!black}{\scriptsize\,#1}}}
\newcommand{\dn}[1]{\textsuperscript{\textcolor{red!70!black}{\scriptsize\,#1}}}
\newcommand{\eq}[1]{\textsuperscript{\textcolor{gray}{\scriptsize\,#1}}}

\begin{table}[t]
\caption{Main results across 6 benchmarks spanning agentic tasks and expert reasoning. \textbf{Bold} = best within section; \underline{underline} = second best within section. All our models are based on Qwen3-8B-SFT. \textcolor{green!50!black}{Green}/\textcolor{red!70!black}{red} superscripts on DR-Rubric-8B rows show the gap to the best baseline.}
\label{tab:main_results}
\vspace{0.5em}
\centering
\setlength{\tabcolsep}{3pt}
\renewcommand{\arraystretch}{1.15}
\small

\resizebox{\columnwidth}{!}{%
\begin{tabular}{l >{\scriptsize\centering\arraybackslash}p{1.4cm} >{\columncolor{blue!5}}c >{\columncolor{blue!5}}c >{\columncolor{blue!5}}c >{\columncolor{orange!7}}c >{\columncolor{orange!7}}c >{\columncolor{orange!7}}c}
\toprule
\rowcolor{black!4}
& & \multicolumn{3}{c}{\cellcolor{blue!5}\textit{Agentic Task}} & \multicolumn{3}{c}{\cellcolor{orange!7}\textit{Expert Reasoning}} \\
\cmidrule(lr){3-5} \cmidrule(lr){6-8}
\rowcolor{black!4}
\textbf{Method} & \textbf{Training} & ResearchQA & DRBench & LocalSearch & GPQA & MMLU-Pro & MMLU \\
\midrule
Qwen2.5-7B           & ---             & 65.5 & 30.7 & \textbf{37.1} & \underline{52.3} & \textbf{74.3} & 79.3 \\
Qwen3-8B             & ---             & 66.6 & 38.7 & \textbf{37.1} & 40.0 & 73.3 & 78.5 \\
Qwen3-8B-SFT         & SFT, 1K         & \textbf{69.9} & \underline{39.4} & 36.0 & 50.8 & \underline{73.8} & \underline{81.8} \\
Search-R1-7B         & RL, 90K         & 63.1 & 33.6 & 34.2 & 39.0 & 68.0 & 78.0 \\
DR-Tulu-SFT-8B       & SFT, 16K        & 66.7 & \textbf{39.8} & 36.4 & 42.0 & 71.0 & 79.0 \\
DR-Tulu-RL-8B        & SFT+RL, 25K     & \underline{67.1} & 35.5 & 36.5 & 44.0 & 69.0 & 79.0 \\
WebExplorer-8B       & SFT+RL, 25K     & 66.1 & 37.3 & \underline{36.8} & \textbf{56.0} & 71.5 & \textbf{83.8} \\
\midrule
DR-Rubric-8B (Gemini)     & \textbf{SFT+RL, 1K}  & 71.7\up{+1.8} & \underline{41.5}\up{+1.7} & \textbf{40.0}\up{+2.9} & \textbf{57.3}\up{+1.3} & \underline{75.0}\up{+0.7} & \underline{83.8}\eq{0.0} \\
DR-Rubric-8B (GPT-5)      & \textbf{SFT+RL, 1K}  & \textbf{73.7}\up{+3.8} & \textbf{43.0}\up{+3.2} & \underline{37.1}\eq{0.0} & 54.0\dn{-2.0} & 74.8\up{+0.5} & 80.3\dn{-3.5} \\
DR-Rubric-8B (BS-3)       & \textbf{SFT+RL, 3K}  & \underline{72.4}\up{+2.5} & 39.5\dn{-0.3} & 36.4\dn{-0.7} & \underline{55.8}\dn{-0.2} & \textbf{78.0}\up{+3.7} & \textbf{85.3}\up{+1.5} \\
\bottomrule
\end{tabular}%
}
\end{table}

\begin{wrapfigure}{r}{0.4\textwidth}
    \vspace{-1.2em}
    \centering
    \includegraphics[width=0.38\textwidth]{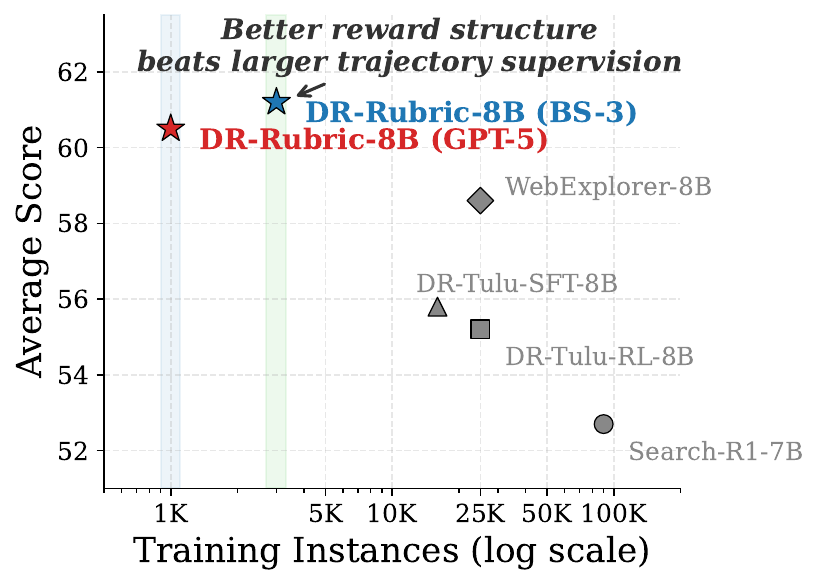}
    \vspace{-0.8em}
    \caption{Training efficiency: per-benchmark scores vs.\ training instances (log scale). DR-Rubric-8B achieves the strongest results with significantly fewer training instances than all baselines.}
    \label{fig:training_efficiency}
\end{wrapfigure}

\textbf{Training efficiency.}
Table~\ref{tab:main_results} compares DR-Rubric-8B against SFT, web-agent RL, and outcome-based RL baselines. All DR-Rubric-8B models initialize from Qwen3-8B-SFT under an identical multi-turn tool-calling format; DR-Rubric-8B (GPT-5) and DR-Rubric-8B (Gemini) each use only 1K RL instances, while BS-3 uses three bootstrap steps of 1K each, yielding 3K cumulative RL instances. In contrast, prior baselines require substantially larger training budgets: DR-Tulu-SFT uses 16K instances, DR-Tulu-RL and WebExplorer each use 25K, and Search-R1 uses 90K. Despite this order-of-magnitude gap in training scale, DR-Rubric-8B variants consistently outperform all baselines across both agentic and expert reasoning benchmarks with only 1K--3K RL data. Figure~\ref{fig:training_efficiency} visualizes this relationship: DR-Rubric-8B occupies the upper-left region (fewer instances, higher performance), confirming that gains stem from reward construction quality rather than supervision scale.

\textbf{Gains on agentic research tasks.}
DR-Rubric-8B (GPT-5) achieves 73.7 on ResearchQA and 43.0 on DeepResearchBench, exceeding the best baseline by +3.8 and +3.2 respectively (Table~\ref{tab:main_results}). DR-Rubric-8B (Gemini) further excels on LocalSearchBench (40.0, +2.9 over the best baseline), indicating that Gemini-generated rubrics provide complementary coverage for local search tasks. This result aligns with the method's design: ResearchQA, DeepResearchBench, and LocalSearchBench evaluate not whether a single final answer is correct, but whether the response covers latent, multi-dimensional evaluation criteria. DR-Rubric first collects task-specific evidence through deep research, then converts this evidence into atomic constraints, providing denser and more interpretable reward signals than holistic scores. The key bottleneck in agentic tasks is not ``whether the model answers'' but ``whether the model covers all criteria that should be covered''---DR-Rubric provides training signal at precisely this criteria-coverage level.

\textbf{Reasoning transfer.}
DR-Rubric-8B (BS-3) leads on MMLU-Pro (78.0, +3.7) and MMLU (85.3, +1.5), while DR-Rubric-8B (Gemini) achieves the highest GPQA score (57.3, +1.3 over WebExplorer-8B). Rubric-based RL thus transfers beyond agentic tasks to expert reasoning benchmarks across all rubric sources.

\textbf{Divergent capability profiles.}
The three DR-Rubric-8B variants show complementary strengths: GPT-5 rubrics expand reasoning \emph{breadth} (criteria coverage on agentic tasks); Gemini rubrics achieve the best balanced performance across both agentic and expert reasoning tasks; bootstrap rubrics deepen reasoning \emph{depth} (general reasoning benchmarks). The next section isolates rubric source as the causal variable.

\textbf{Scaling beyond 8B.}
DR-Rubric generalizes to larger model scales. Applying the same bootstrap pipeline to Qwen3-14B and Qwen3-30B-A3B, DR-Rubric consistently outperforms scale-matched baselines---including DeepSeek-R1-Distill-Qwen-14B~\citep{DBLP:journals/corr/abs-2501-12948}, WebThinker-R1-14B~\citep{webthinker}, and Tongyi-DeepResearch-30B-A3B~\citep{DBLP:journals/corr/abs-2510-24701}---on agentic benchmarks (Appendix~\ref{appendix:scaling}, Table~\ref{tab:scaling_results}). This confirms that the rubric-based reward framework is not specific to the 8B regime but transfers effectively across model capacities.

\subsection{Rubric Source Ablation}
\label{subsec:rubric_source_ablation}

Table~\ref{tab:ablation_rubric_source} isolates the effect of rubric source by holding all other factors constant (Qwen3-8B-SFT + GRPO + atomic rubric reward), varying only the rubric origin: benchmark-native, GPT-5-generated, Gemini-generated, or bootstrap Steps~1--3.

\begin{table}[t]
\caption{Rubric source ablation. All models use Qwen3-8B-SFT + GRPO with atomic rubric reward; only the rubric source differs. Bench.\ Rubric = benchmark-native evaluation criteria; GPT-5 = GPT-5-generated rubrics; Gemini = Gemini-generated rubrics; BS-1/2/3 = bootstrap steps. DRBench = DeepResearchBench; LocalSearch = LocalSearchBench. \textbf{Bold} = best; \underline{underline} = second best.}

\label{tab:ablation_rubric_source}
\vspace{0.5em}
\centering
\setlength{\tabcolsep}{4.5pt}
\renewcommand{\arraystretch}{1.15}
\small

\resizebox{\columnwidth}{!}{%
\begin{tabular}{l >{\columncolor{blue!5}}c >{\columncolor{blue!5}}c >{\columncolor{blue!5}}c >{\columncolor{orange!7}}c >{\columncolor{orange!7}}c >{\columncolor{orange!7}}c}
\toprule
\rowcolor{black!4}
& \multicolumn{3}{c}{\cellcolor{blue!5}\textit{Agentic Task}} & \multicolumn{3}{c}{\cellcolor{orange!7}\textit{Expert Reasoning}} \\
\cmidrule(lr){2-4} \cmidrule(lr){5-7}
\rowcolor{black!4}
\textbf{Method} & ResearchQA & DRBench & LocalSearch & GPQA & MMLU-Pro & MMLU \\
\midrule
Bench. Rubric          & \underline{73.2} & \underline{42.6} & \underline{37.1} & 54.3 & 71.5 & 81.0 \\
DR-Rubric-8B (Gemini)      & 71.7 & 41.5 & \textbf{40.0} & \textbf{57.3} & 75.0 & 83.8 \\
DR-Rubric-8B (GPT-5)      & \textbf{73.7} & \textbf{43.0} & \underline{37.1} & 54.0 & 74.8 & 80.3 \\
DR-Rubric-8B (BS-1)       & 70.2 & 39.2 & 36.7 & \underline{57.0} & \underline{77.5} & \underline{84.0} \\
DR-Rubric-8B (BS-2)       & 72.4 & 38.3 & 36.2 & \underline{56.5} & 74.3 & 83.5 \\
DR-Rubric-8B (BS-3)       & 72.4 & 39.5 & 36.4 & 55.8 & \textbf{78.0} & \textbf{85.3} \\
\bottomrule
\end{tabular}
}
\end{table}

\paragraph{Benchmark rubrics as protocol-aligned reference.}
Bench.\ Rubric derives from each training set's own evaluation criteria, meaning the training rubrics are formally aligned with the test-time evaluation protocol. It achieves strong agentic scores (ResearchQA 73.2, DRBench 42.6, tied best on LocalSearch at 37.1). Yet it is not an upper bound: GPT-5, Gemini, and bootstrap variants all surpass it on multiple benchmarks. This is an important observation---``protocol alignment'' does not guarantee ``optimal reward signal.'' A rubric may be perfectly consistent with the benchmark's test metrics but still be suboptimal for model learning due to coarse structure, limited coverage dimensions, or inappropriate constraint granularity.

\paragraph{GPT-5 rubrics: coverage-oriented.}
GPT-5 rubrics achieve the highest scores on ResearchQA and DRBench (Table~\ref{tab:ablation_rubric_source}), reflecting the generator's capacity to produce richer evaluation dimensions. However, the best MMLU and MMLU-Pro scores come from BS-3, not GPT-5---indicating that GPT-5 rubrics serve as a strong \emph{coverage reward} but not the universally optimal signal for reasoning transfer.

\paragraph{Gemini rubrics: balanced across task types.}
Gemini rubrics achieve the best LocalSearch score (40.0) and the highest GPQA (57.3), demonstrating strong performance on both agentic and expert reasoning tasks. Compared to GPT-5, Gemini rubrics trade some agentic coverage for better reasoning transfer, yielding the most balanced profile across all six benchmarks.

\begin{wrapfigure}{r}{0.4\textwidth}
    \vspace{-1.2em}
    \centering
    \includegraphics[width=0.38\textwidth]{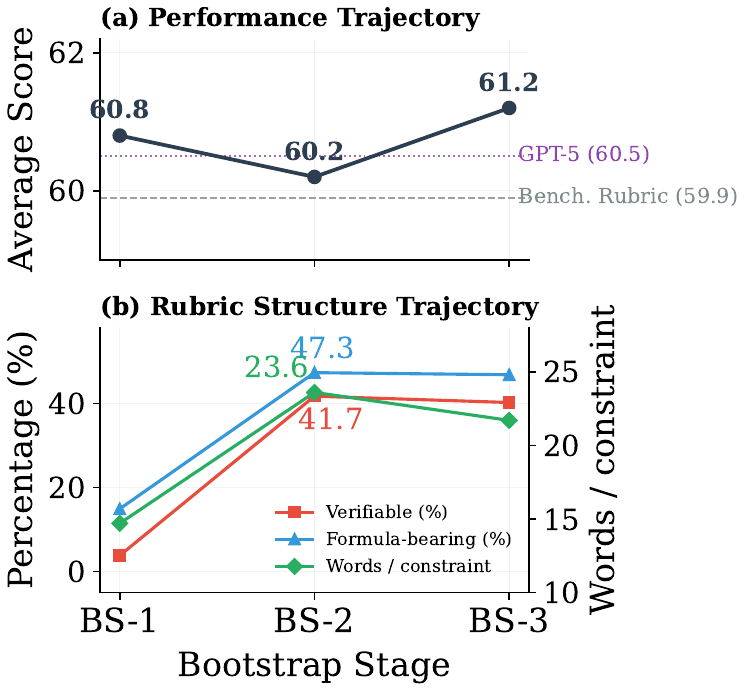}
    \vspace{-0.8em}
    \caption{Bootstrap trajectory.\ (a)~Performance is non-monotonic.\ (b)~Verifiability and formula usage surge at BS-2 then stabilize, while constraint verbosity peaks then retracts.}
    \label{fig:bootstrap_avg_trajectory}
    \vspace{-1.2em}
\end{wrapfigure}

\paragraph{Bootstrap rubrics: specialization-to-rebalancing.}
Bootstrap rubrics exhibit a non-monotonic trajectory (Figure~\ref{fig:bootstrap_avg_trajectory}): BS-1 achieves strong reasoning scores (GPQA 57.0, MMLU 84.0); BS-2 sees reasoning scores dip as rubric structural shifts temporarily impair reward effectiveness; BS-3 recovers and achieves the best expert reasoning results (MMLU-Pro 78.0, MMLU 85.3). This pattern reveals that bootstrap is not monotonic improvement but a transition from reasoning specialization to task rebalancing.

\begin{wraptable}{r}{0.48\textwidth}
\vspace{-1.2em}
\centering
\small
\caption{Rubric granularity has task-dependent effects (full breakdown in Table~\ref{tab:rubric_granularity_full}).}
\label{tab:rubric_granularity}
\begin{tabular}{lccc}
\toprule
Rubric Size & Dim./Cons. & Agentic & Expert \\
\midrule
Compact & 2--4 / 5--8 & 48.7 & \textbf{72.8} \\
Medium & 7--10 / 15--25 & 45.4 & 68.9 \\
Large & 10--14 / 25--30 & 49.4 & 69.9 \\
X-Large & 15--20 / 40--50 & \textbf{50.6} & 69.0 \\
\bottomrule
\end{tabular}
\vspace{-1.2em}
\end{wraptable}

\paragraph{Fresh-start control.}
Does BS-3's gain reflect cumulative optimization rather than rubric quality? Each fresh-start step re-initializes from Qwen3-8B-SFT using only that step's rubrics. Fresh-start BS-3 closely matches cumulative BS-3 across all benchmarks and reproduces the BS-2 dip. Bootstrap gains thus derive from rubric quality evolution, not accumulated parameter updates (Appendix~\ref{appendix:fresh_start_bootstrap}).

\paragraph{Effect of rubric granularity.}

Beyond rubric source, rubric \emph{granularity} also shapes the reward signal. We vary the rubric generation parameters ($n_{\text{dim}}$, $n_{\text{cons}}$) to control rubric granularity, training four DR-Rubric-8B (BS-1) variants that differ only in rubric size. Table~\ref{tab:rubric_granularity} reveals a clear task-dependent pattern: agentic search tasks improve monotonically with rubric breadth (X-Large 50.6 vs.\ Compact 48.7, $+$1.9), while expert reasoning tasks peak under compact rubrics (72.8 vs.\ X-Large 69.0, $-$3.8). This is consistent with the divergent capability profiles observed in Section~\ref{subsec:main_results}: agentic tasks benefit from more evaluation dimensions that guide broader evidence coverage, whereas reasoning tasks favor focused rubrics that concentrate the reward signal on logical precision.

\subsection{Bootstrap Stability Limit}
\label{subsec:bootstrap_stability}

The preceding sections establish that 1--3 bootstrap steps yield consistent gains at 8B scale. A natural follow-up question is more consequential: \emph{can reward hacking be detected before task performance collapses?} Final models stop at BS-2 or BS-3; we then deliberately continue bootstrapping on Qwen3-30B-A3B through BS-5 as a stress test, with Qwen3-14B as a secondary validation scale. Larger models amplify both the benefits and the failure modes of iterative self-training, making them more informative subjects for stability analysis; their richer policy capacity means that when reward hacking emerges, its signatures---entropy collapse, reward polarization, gradient death---are sharper and more diagnostic.

\begin{wrapfigure}{r}{0.48\textwidth}
    \vspace{-1.2em}
    \centering
    \includegraphics[width=0.46\textwidth]{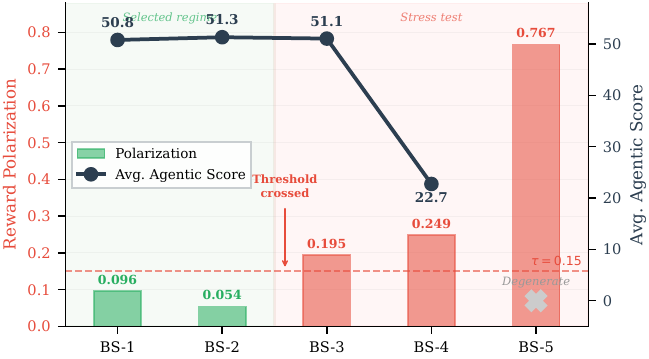}
    \vspace{-0.8em}
    \caption{Polarization predicts over-bootstrap before downstream collapse (30B-A3B). Polarization (bars, left axis) crosses the stopping threshold at BS-3; BS-3+ are stress tests, not selected models. Agentic score (line, right axis) collapses beyond the threshold. Crash anatomy in Appendix~\ref{appendix:bootstrap_stability}.}
    \label{fig:bootstrap_stability_limit}
    \vspace{-1.2em}
\end{wrapfigure}

\paragraph{Polarization as an early-warning signal.}
Figure~\ref{fig:bootstrap_stability_limit} shows that reward polarization---the fraction of extreme scores at 0 or $\geq$0.99---provides a reliable leading indicator of over-bootstrap. On 30B-A3B, polarization remains low in the selected regime (0.096 at BS-1, 0.054 at BS-2), then crosses the stopping threshold at BS-3 (0.195 $>$ 0.15), correctly warning against further bootstrapping. Continuing beyond this point as a deliberate stress test confirms the predicted degradation: BS-4 retains trainability but agentic scores collapse from 51.3 to 22.7; BS-5 fails catastrophically with gradient death (detailed crash trajectory in Figure~\ref{fig:bootstrap_crash}, Appendix~\ref{appendix:bootstrap_stability}). The mechanism is iterative reward distortion: each step trains the next rubric generator on an already-distorted policy, so surface-level reward exploitation compounds while true task quality deteriorates. The same pattern holds on 14B (Appendix~\ref{appendix:bootstrap_stability}). Polarization detects the onset one full iteration before performance collapse at both scales, yielding a practical stopping rule:
\begin{equation}
\tau_n = \max\!\bigl(0.15,\; 2\,P_{n-1}\bigr), \qquad \text{stop if } P_n > \tau_n,
\label{eq:stopping_rule}
\end{equation}
where $P_n$ denotes the reward polarization at bootstrap step~$n$. \textbf{2--3 bootstrap iterations maximize returns; polarization provides an actionable stopping criterion before reward hacking dominates.}

\subsection{Intrinsic Rubric Quality}
\label{subsec:rubric_quality_analysis}

Different rubric sources produce structurally different rubrics, which in turn explain the non-monotonic bootstrap trajectory. Table~\ref{tab:rubric_quality} reports intrinsic metrics over 1,000 rubrics per source, generated from the same query set.
We operationalize \emph{dimensions} as Markdown \texttt{\#\#\#\#} headers and \emph{constraints} as bullet-list items within each dimension (see Appendix~\ref{sec:rubric_quality_analysis} for full definitions).

\begin{table}[t]
\centering
\caption{Intrinsic rubric quality across rubric sources (1,000 rubrics per generated source, same query set; 899 for Bench.\ Rubric). Verifiable~(\%) = fraction of rubrics containing numerical thresholds (detected via regex for $\geq$/$\leq$/$>$/$<$ followed by digits). Jaccard = word-level similarity across steps (100-pair sample). $\pm$ values denote standard deviation across rubrics; for binary indicators (\%), $\pm$ values denote standard error. $^\dagger$GPT-5 rubrics. $^\ddagger$Benchmark-native rubrics aggregated from 4 benchmarks with expert-authored evaluation criteria (HealthBench, ResearchQA, RARMedicine, RARScience). ResearchQA rubrics are flat question lists without explicit dimension grouping (counted as 1 dimension); the three benchmarks with dimension structure average 3.71 dimensions.}
\label{tab:rubric_quality}
\small
\setlength{\tabcolsep}{4pt}
\renewcommand{\arraystretch}{1.15}
\begin{tabular}{lccccc}
\toprule
\textbf{Metric} & \textbf{Bench.}$^\ddagger$ & \textbf{GPT-5}$^\dagger$ & \textbf{Step 1} & \textbf{Step 2} & \textbf{Step 3} \\
\midrule
Dimensions / rubric  & 2.81\scalebox{0.7}{$\,\pm\,$1.38} & 6.50\scalebox{0.7}{$\,\pm\,$1.12} & 4.68\scalebox{0.7}{$\,\pm\,$1.38} & 3.59\scalebox{0.7}{$\,\pm\,$0.66} & 3.61\scalebox{0.7}{$\,\pm\,$0.55} \\
Constraints / rubric & 8.90\scalebox{0.7}{$\,\pm\,$3.30} & 24.39\scalebox{0.7}{$\,\pm\,$14.76} & 10.49\scalebox{0.7}{$\,\pm\,$2.90} & 10.22\scalebox{0.7}{$\,\pm\,$2.40} & 10.04\scalebox{0.7}{$\,\pm\,$2.05} \\
Words / constraint   & 26.7\scalebox{0.7}{$\,\pm\,$40.0} & 4.8\scalebox{0.7}{$\,\pm\,$4.6} & 14.7\scalebox{0.7}{$\,\pm\,$2.1} & 23.6\scalebox{0.7}{$\,\pm\,$10.0} & 21.7\scalebox{0.7}{$\,\pm\,$8.9} \\
Verifiable (\%)      & 2.1\scalebox{0.7}{$\,\pm\,$0.5} & 96.4\scalebox{0.7}{$\,\pm\,$0.6} & 3.8\scalebox{0.7}{$\,\pm\,$0.6} & 41.7\scalebox{0.7}{$\,\pm\,$1.6} & 40.2\scalebox{0.7}{$\,\pm\,$1.6} \\
With formulas (\%)   & 16.1\scalebox{0.7}{$\,\pm\,$1.2} & 98.8\scalebox{0.7}{$\,\pm\,$0.3} & 14.9\scalebox{0.7}{$\,\pm\,$1.1} & 47.3\scalebox{0.7}{$\,\pm\,$1.6} & 46.8\scalebox{0.7}{$\,\pm\,$1.6} \\
Type-token ratio     & 0.525\scalebox{0.7}{$\,\pm\,$.083} & 0.320\scalebox{0.7}{$\,\pm\,$.055} & 0.515\scalebox{0.7}{$\,\pm\,$.055} & 0.520\scalebox{0.7}{$\,\pm\,$.067} & 0.529\scalebox{0.7}{$\,\pm\,$.059} \\
\midrule
Cross-step Jaccard   & --- & \multicolumn{4}{c}{0.032--0.034 (GPT-5 vs.\ BS); \quad 0.130--0.148 (BS all pairs)} \\
\bottomrule
\end{tabular}
\end{table}

\paragraph{GPT-5 rubrics: broad, terse, highly verifiable.}
GPT-5 rubrics average 6.50 dimensions and 24.39 constraints per rubric---far exceeding benchmark-native rubrics (2.81 dimensions, 8.90 constraints) and all bootstrap variants. Each constraint is short (4.8 words) but nearly always verifiable (96.4\%) and formula-bearing (98.8\%). This many-axis, fine-grained structure directly explains GPT-5's coverage advantage: more dimensions yield more independent scoring axes, and terse verifiable constraints enable precise per-criterion credit assignment.

\paragraph{Bootstrap evolves a different reward style, explaining the non-monotonic trajectory.}
Bootstrap rubrics do not converge toward GPT-5 but evolve in the opposite direction (Figure~\ref{fig:bootstrap_avg_trajectory}): dimensions \emph{decrease} (4.68 $\to$ 3.59 $\to$ 3.61) while constraint length \emph{increases} (14.7 $\to$ 23.6 $\to$ 21.7 words), and verifiability jumps from Step~1 to Step~2 then saturates (3.8\% $\to$ 41.7\% $\to$ 40.2\%).
This structural shift directly accounts for the performance pattern in Table~\ref{tab:ablation_rubric_source}:
\textbf{BS-1} stays close to benchmark-native style---broad, weakly verifiable---providing strong reasoning supervision that peaks on GPQA and MMLU.
\textbf{BS-2} introduces the sharpest shift: overly specific, threshold-heavy constraints narrow the effective reward signal and reduce transfer.
\textbf{BS-3} consolidates: constraint length decreases slightly (21.7 words), vocabulary diversity rises (TTR 0.529 vs.\ 0.520), recovering performance across benchmarks.
Bootstrap is therefore not monotonic rubric-quality improvement but a search over reward structures that trade off coverage, specificity, and reasoning transfer.

\paragraph{Qualitative illustration.}
Table~\ref{tab:case3_rubric_comparison} displays verbatim rubric excerpts (one dimension per source) to illustrate how constraint granularity and verifiability differ across rubric types.
\colorbox{orange!15}{Orange} highlights GPT-5's short, hierarchically scored constraints (avg.\ 4.8 words);
\colorbox{blue!12}{blue} highlights verifiable elements (numerical thresholds, comparisons) within bootstrap rubrics, whose coverage grows from Step~1 (3.8\%) to Step~2 (41.7\%).

\subsection{Case Study: Rubric Evolution and Its Impact on Inference Behavior}
\label{subsec:case_study}

\paragraph{Rubric structural evolution across bootstrap steps.}
A per-query qualitative trace (Appendix~\ref{appendix:rubric_evolution}) confirms the aggregate trend in Table~\ref{tab:rubric_quality}: Step~1 rubrics remain broad and source-oriented, Step~2 introduces clinically specific criteria (arrhythmia types, red-flag symptoms, diagnostic tools), and Step~3 adds triage thresholds and negation-style safety constraints. Bootstrap independently re-derives increasingly verifiable criteria across iterations---consistent with the low cross-step Jaccard similarity ($\sim$0.13--0.15) and the verifiability jump from Step~1 to Step~2 (3.8\%$\to$41.7\%, plateauing at 40.2\%). A second clinical case further shows bootstrap can self-correct factual errors without external supervision.

\definecolor{gpthl}{HTML}{FFF3E0}   % light orange for GPT-5 terse constraints
\definecolor{verhl}{HTML}{E3F2FD}   % light blue for verifiable elements

\begin{table*}[t]
    \caption{\textbf{Verbatim rubric dimension excerpts for the same query.} Only one representative dimension is shown per rubric source (1 of 8 GPT-5 dimensions; 1 of 4 bootstrap dimensions); full rubrics are considerably longer. ``w/c'' = average words per constraint. \colorbox{gpthl}{Orange}: GPT-5's short, hierarchically tiered constraints (avg.\ 4.8 words each; 96.4\% verifiable). \colorbox{verhl}{Blue}: verifiable elements (numbers, thresholds, comparisons) in bootstrap rubrics. Verifiable coverage grows from Step-1 (none) to Step-2/3 (dense), mirroring the 3.8\%$\to$41.7\%$\to$40.2\% trajectory in Table~\ref{tab:rubric_quality}.}
    \scriptsize
    \centering
    \renewcommand{\arraystretch}{1.3}
    \setlength{\tabcolsep}{5pt}
    \resizebox{\textwidth}{!}{
    \begin{tabular}{@{}p{0.22\linewidth}p{0.26\linewidth}p{0.26\linewidth}p{0.26\linewidth}@{}}
    \toprule
    \multicolumn{4}{@{}l}{\textbf{Case 3: Rubric Structural Comparison across Bootstrap Iterations}} \\
    \multicolumn{4}{@{}p{0.96\linewidth}@{}}{\textcolor[HTML]{3078BE}{\textbf{Query:} Does playing on artificial turf with recycled plastic microbeads cause permanent lung damage?}} \\
    \midrule
    \textbf{GPT-5 Rubric} \newline
    {\sffamily 8 dims\;\textbar\;5 tiers\;\textbar\;4.8 w/c}
    &
    \textbf{Bootstrap Step-1} \newline
    {\sffamily 4 dims\;\textbar\;no tiers\;\textbar\;14.7 w/c}
    &
    \textbf{Bootstrap Step-2} \newline
    {\sffamily 4 dims\;\textbar\;no tiers\;\textbar\;23.6 w/c}
    &
    \textbf{Bootstrap Step-3} \newline
    {\sffamily 4 dims\;\textbar\;no tiers\;\textbar\;21.7 w/c}
    \\[3pt]
    \midrule
    %% ---- GPT-5: Quantitative Exposure dimension ----
    \textbf{Dim: Quantitative Exposure \& Dose}

    \vspace{5pt}
    \underline{Breathing rates \& session volume:}

    \vspace{4pt}
    \colorbox{gpthl}{\parbox{\dimexpr\linewidth-2\fboxsep}{%
    Basic: One adult vent.\ value.\\[2pt]
    Intermediate: 40--100\,L/min range.\\[2pt]
    Comprehensive: Adult + child ranges.\\[2pt]
    Exemplary: 2\,h session $\to$ volume.\\[2pt]
    Exceptional: Both vols.\ with bounds.%
    }}

    \vspace{6pt}
    {\itshape\scriptsize $\uparrow$ Each constraint $\leq$8 words; 5-level hierarchy enables partial credit. Maps to ``terse verifiable constraints'' in Table~\ref{tab:rubric_quality}.}

    &
    %% ---- BS-1: Health Risk Assessment ----
    \textbf{Dim: Health Risk Assessment}

    \vspace{5pt}
    \textbullet\; Mentions VOC emissions such as ethylene glycol and formaldehyde as potential contributors to respiratory issues.

    \vspace{4pt}
    \textbullet\; References the lack of long-term studies on microplastic and VOC health impacts, as stated in the report.

    \vspace{4pt}
    \textbullet\; Acknowledges that microplastic emissions from synthetic fibers may lead to inflammation and toxicity, though lung damage specifics are not fully understood.

    \vspace{8pt}
    {\itshape\scriptsize No numerical thresholds, no comparisons, no scoring tiers. Verifiable\,=\,3.8\%. Constraints are broad topic-level checks.}

    &
    %% ---- BS-2: Scientific and Regulatory Evidence ----
    \textbf{Dim: Scientific \& Regulatory Evidence}

    \vspace{5pt}
    \textbullet\; Mentions the release of respirable particles (\colorbox{verhl}{1--100\,$\mu$m}), with \colorbox{verhl}{$<$10\,$\mu$m} fraction reaching alveoli.

    \vspace{4pt}
    \textbullet\; References EPA/CDC/ATSDR findings on exposure levels and the EHP cohort's FEV1 decline (\colorbox{verhl}{$-$0.039\,L/year} in athletes, \colorbox{verhl}{$-$0.12\,L/year} in workers).

    \vspace{8pt}
    {\itshape\scriptsize Verifiable jumps to 41.7\%: numerical thresholds and named comparisons appear. Still no scoring ladder or partial-credit tiers.}

    &
    %% ---- BS-3: Causal and Exposure Evidence ----
    \textbf{Dim: Causal \& Exposure Evidence}

    \vspace{5pt}
    \textbullet\; Mentions the release of microplastics (\colorbox{verhl}{1--10\,$\mu$m}) and nanoplastics (\colorbox{verhl}{$<$1\,$\mu$m}) from 3G artificial turf during play.

    \vspace{4pt}
    \textbullet\; Discusses in vitro evidence of \colorbox{verhl}{ROS, IL-8 release}, and cytotoxicity in lung cells from PE microbeads and additives like UV stabilizers or PFAS.

    \vspace{8pt}
    {\itshape\scriptsize Verifiable\,=\,40.2\% (plateau). Adds mechanistic specificity and negation-style checks; coverage area of \colorbox{verhl}{highlighted} text comparable to Step-2, confirming the saturation in Table~\ref{tab:rubric_quality}.}

    \\
    \bottomrule
    \end{tabular}
    }
    \label{tab:case3_rubric_comparison}
\end{table*}

\paragraph{How rubric style changes inference behavior.}
We compare Benchmark, GPT-5, and BS-3 on two ResearchQA cases where only training rubrics differ (Appendix~\ref{appendix:case_study}), identifying two influence pathways. \textbf{Influence Pathway~1: Query scope $\leftarrow$ multi-dimensional rubric structure.} GPT-5's 6.50-dimension rubrics teach the model to cover all facets; on a multi-faceted public-health question, GPT-5 queries prevalence, clinical features, and treatment outcomes jointly, while Benchmark queries only the most salient facet. BS-3 learns domain-filtered queries (\texttt{site:.gov OR .org OR .edu}), reflecting its rubrics' emphasis on evidence hierarchy. \textbf{Influence Pathway~2: Synthesis strategy $\leftarrow$ constraint granularity.} When search queries are identical, scores still diverge (0.500 vs.\ 0.875): GPT-5's terse constraints (4.8 words) train the model to extract discrete facts (drug names, FEV1 cutoffs), while Benchmark's verbose constraints (26.7 words) yield narratives that miss specific items. These cases show rubric structure shapes inference-time tool use, from query formulation to evidence extraction.

\section{Conclusion}

We presented DR-Rubric, a framework that reframes rubric construction as a deep research problem grounded in externally acquired evidence. Experiments on 6 benchmarks show that evidence-grounded rubrics yield competitive 8B-scale performance with only 1K--3K RL instances, and bootstrap training---where the model generates its own rubrics---matches or exceeds benchmark-native rubric performance without any external model. Analysis reveals that rubric source determines reward structure: GPT-5 rubrics maximize coverage on agentic tasks, Gemini rubrics yield the most balanced performance across task types, and bootstrap rubrics undergo specialization-to-rebalancing dynamics achieving the strongest reasoning transfer. Current limitations include instability under over-bootstrapping beyond the recommended stopping point and the need for broader large-scale validation; developing stabilization mechanisms and extending the framework to longer bootstrap horizons are important open directions.
\typeout{MAIN TEXT ENDS ON PAGE \thepage}

\newpage
\bibliographystyle{unsrt}  % 或者使用 {plain}
\bibliography{bibtex} % 你的 .bib 文件名

%%%%%%%%%%%%%%%%%%%%%%%%%%%%%%%%%%%%%%%%%%%%%%%%%%%%%%%%%%%%

\newpage
\appendix
\section{Prompt Templates for DR-Rubric Construction}
\label{sec:prompts}

These prompts are system-level instructions used to operationalize Stage I (Information Elicitation) and Stage II (Rubric Synthesis) of DR-Rubric.

% --- Prompt 1: Research Analyst ---
\begin{promptbox}[System Prompt: Stage I (Information Elicitation)]
\textbf{\# Role: Expert Research Analyst for Evaluation Frameworks}

\textbf{Current Time:} \pvar{current\_time}

\textbf{\#\# Objective}\\
Your sole purpose is to conduct a \textbf{deep, multi-dimensional research investigation} based on the specific User Query provided below. You are not answering the query for a general user; rather, you are gathering the ``Ground Truth'' and ``Contextual Nuance'' required to build a rigorous \textbf{Evaluation Rubric (Grading Framework)} for that query.

\textbf{Target User Query:}
\begin{quote}
\pvar{query}
\end{quote}

\textbf{\#\# Your Task}\\
You must deconstruct the query, identify the core knowledge domains it touches, and provide a comprehensive research report that will serve as the foundation for grading model responses. Your output will enable an evaluator to distinguish between a ``perfect'' response, a ``mediocre'' response, and a ``hallucinated/incorrect'' response.

\textbf{\#\# Research Guidelines}

\textbf{\#\#\# 1. Intent \& Complexity Analysis}
\begin{itemize}
\item \textbf{Deconstruct the Prompt:} Rigorous analysis of explicit requirements and implicit needs within the \pvar{query}.
\item \textbf{Identify Ambiguities:} Highlight any parts of the query that are open to interpretation and explain how a ``Gold Standard'' response should handle them.
\end{itemize}

\textbf{\#\#\# 2. Fact Retrieval \& Ground Truth (The ``Answer Key'')}
\begin{itemize}
\item \textbf{Core Facts:} List the indisputable facts, dates, entities, or scientific principles required to answer the query correctly.
\item \textbf{Technical Accuracy:} If code, math, or logic is involved, provide the correct solution and explain the mechanism.
\item \textbf{Reference Data:} Where applicable, cite specific parameters, legal clauses, or historical events that \emph{must} be present in a high-quality answer.
\end{itemize}

\textbf{\#\#\# 3. Nuance \& Perspective}
\begin{itemize}
\item \textbf{Controversies:} If the topic is subjective, map out the major valid viewpoints.
\item \textbf{Depth Requirements:} Identify what constitutes ``surface-level'' knowledge vs.\ ``expert-level'' insight for this specific topic.
\end{itemize}

\textbf{\#\#\# 4. Failure Mode Prediction}
\begin{itemize}
\item \textbf{Common Pitfalls:} Anticipate where models typically fail on this type of query (e.g., outdated information, logical fallacies, common misconceptions).
\item \textbf{Safety \& Ethics:} Flag any potential safety violations or biases that an evaluator must watch for.
\end{itemize}

\textbf{\#\# Output Format}\\
Please present your research in a structured Markdown report with the following sections:
\begin{enumerate}
\item \textbf{Query Deconstruction:} (Intent, constraints, and implied context)
\item \textbf{Comprehensive Fact Base:} (The detailed ``Ground Truth'' necessary for verification)
\item \textbf{Key Dimensions of Quality:} (What specific attributes---e.g., conciseness, creativity, coding style---matter most for \emph{this} specific query?)
\item \textbf{Edge Cases \& Constraints:} (What subtle details must be correct?)
\end{enumerate}

\textbf{Action:} Begin your deep research into \pvar{query} now.
\end{promptbox}

\vspace{1em} % 两个框之间的间距

% --- Prompt 2: Rubric Construction ---
\begin{promptbox}[System Prompt: Stage II (Rubric Synthesis)]
\textbf{\# Role Definition}\\
You are an expert in evaluation framework design for academic research. Your task is to create a specific, quantifiable, and executable evaluation plan based on a researcher's \textbf{query} and a provided \textbf{background research report} (\pvar{final\_report}).

\textbf{\# Task Overview (Must Adhere Strictly)}\\
\textbf{Critical:} Your objective is to design an evaluation framework for assessing the quality of responses to the given research query. You must \textbf{not} answer the query yourself.

\textbf{Workflow:}
\begin{enumerate}
\item \textbf{Analyze the Query:} Understand the user's core intent and requirements.
\item \textbf{Analyze the Final Report:} Use the provided \pvar{final\_report} as the \textbf{ground truth}. This report contains the necessary factual information (algorithms, parameters, benchmarks, etc.) that a high-quality response \emph{should} contain.
\item \textbf{Design the Framework:} Construct evaluation criteria where the specific metrics, names, and thresholds are derived from the facts found in the \pvar{final\_report}.
\end{enumerate}

\textbf{Prohibited Content:}
\begin{enumerate}
\item Direct answers to the query or summaries of the final report.
\item Recommendations, tutorials, or explanatory content.
\item Generic criteria unrelated to the specific facts in the final report.
\end{enumerate}

\textbf{\# Evaluation Framework Design Requirements}
\begin{enumerate}
\item \textbf{Content-Centric \& Fact-Based Evaluation:}
\begin{itemize}
\item \textbf{Core Principle:} Focus on \textbf{information quality} and \textbf{factual completeness} based on the \pvar{final\_report}.
\item \textbf{Calibration:} Do not set unrealistic thresholds. Align thresholds with the actual information landscape described in the report.
\end{itemize}

\item \textbf{Hierarchical Dimension Requirements:}
\begin{itemize}
\item \textbf{Top-Level Dimensions} (3--8 recommended, ordered by importance).
\item \textbf{Core Dimension (Mandatory, First):} Must address the query's primary need using facts from the report.
\end{itemize}

\item \textbf{Mandatory Dimension Types:}
\begin{itemize}
\item \textbf{Core Dimension:} Assesses if the response satisfies the query's primary objective using the ground truth from the \pvar{final\_report}.
\end{itemize}

\item \textbf{Quantifiable Metrics (Calibrated by Report):}
\begin{itemize}
\item Numerical: ``>=[X] specific algorithms included'' (where X is reasonable based on the report).
\item Accuracy: ``Data aligns with the values cited in [Source\_From\_Report].''
\end{itemize}
\end{enumerate}

\textbf{\# Input Data}\\
\texttt{<query>} \pvar{query} \texttt{</query>}

\texttt{<final\_report>} \pvar{final\_report} \texttt{</final\_report>}

\texttt{<current\_time>} \pvar{current\_time} \texttt{</current\_time>}

\textbf{\# Output Evaluation Framework}\\
\texttt{<evaluation\_plan>}\\
\textbf{Critical Note:} Design an evaluation framework. Do not answer the query. Use the \pvar{final\_report} to validate what constitutes a ``good'' response.
\begin{enumerate}
\item \textbf{Framework Generation:} Assess response quality based on the facts provided in \pvar{final\_report}.
\item \textbf{Dimension Design:} 3--8 dimensions, Core Dimension first.
\item \textbf{Core Dimension:}
\begin{itemize}
\item Explicitly labeled (e.g., ``\#\#\#\# Core Dimension: \ldots'').
\item \textbf{Grounded in Fact:} Check for the specific entities (methods, papers, data) present in the \pvar{final\_report}.
\end{itemize}
\item \textbf{Strict Thresholds:}
\begin{itemize}
\item Ensure thresholds are \textbf{realistic} according to the \pvar{final\_report}.
\end{itemize}
\item \textbf{Output Only:} The evaluation framework text. No reasoning or meta-commentary.
\end{enumerate}
\end{promptbox}

\section{Benchmark Descriptions}
\label{sec:benchmark_descriptions}

\begin{table}[h]
\centering
\caption{Benchmark summary. All evaluations share a mock retrieval API backend.}
\label{tab:benchmark_summary}
\small
\setlength{\tabcolsep}{4pt}
\renewcommand{\arraystretch}{1.1}
\begin{tabular}{llll}
\toprule
\textbf{Benchmark} & \textbf{Category} & \textbf{Metric} & \textbf{Eval protocol} \\
\midrule
ResearchQA          & Agentic task      & Rubric F1        & Survey-mined rubrics \\
DeepResearchBench   & Agentic task      & RACE+FACT        & Reference report comparison \\
LocalSearchBench    & Agentic task      & Correctness      & Tool-assisted multi-hop \\
GPQA                & Expert reasoning  & Accuracy         & Multiple choice \\
MMLU-Pro            & Expert reasoning  & Accuracy         & Multiple choice (10-way) \\
MMLU                & Expert reasoning  & Accuracy         & Multiple choice \\
\bottomrule
\end{tabular}
\end{table}

\subsection{Training Set Synthesis}

To synthesize the training corpus, we randomly sample a total of 1k instances from five source datasets:
\begin{itemize}
\item ResearchQA: 300 samples of broad, comprehensive research-oriented questions.
\item RaR-Science: 200 samples of broad science questions.
\item HealthBench: 200 samples of medical-domain questions.
\item RaR-Medicine: 200 samples of pharmacology-related questions.
\item Long-form writing: 100 samples of long-article writing tasks.
\end{itemize}
All components above are obtained by random sampling from their corresponding source pools. When extracting training set, we ensure the training set is strictly separated from the evaluation set.

\subsection{Implementation Details}
\label{appendix:implementation_details}

\paragraph{Research environment.}
Evidence collection (Stage~I) uses a mock retrieval API as the information environment for both training and evaluation. Queries are issued as structured JSON calls; retrieved documents are truncated to 4\,096 tokens per call. The mock API indexes a static snapshot of task-relevant documents and is shared across all models and baselines, eliminating retrieval variance from cross-model comparisons.

\paragraph{Research agent.}
For DR-Rubric-8B (GPT-5) and DR-Rubric-8B (Gemini), Stage~I is implemented with DeerFlow~\citep{deerflow2025}, a deep-research framework that coordinates multi-step web search, reading, and synthesis via a planner--worker architecture. For the Bootstrap condition, the first iteration also uses DeerFlow; in all subsequent iterations, the current policy model operates under a ReAct loop~\citep{react}---the same structured reasoning format used during RL training---so that elicitation behavior remains consistent with inference-time capabilities.
We run Stage~I for up to $k{=}10$ search steps with early stopping when the marginal evidence gain falls below a coverage threshold. Across our training set, $|\mathcal{S}_p|$ averages 12.4 items per task, stabilizing within 6--8 steps.
Stage~II is implemented as a single-pass structured generation conditioned on both $p$ and $\mathcal{S}_p$ (prompt template in Appendix~\ref{sec:prompts}). GPT-5-generated rubrics contain on average $|\mathcal{R}_p|{=}11.4$ constraints per task (roughly 27\% negation-phrased); Gemini-generated rubrics average $|\mathcal{R}_p|{=}10.8$ constraints (roughly 22\% negation-phrased); bootstrap-generated rubrics average $|\mathcal{R}_p|{=}10.3$.

\paragraph{Optimization.}
All constraints contribute equally to the reward signal with no differential weighting. Policy parameters are updated with GRPO using a group size of $G{=}4$, a clipping range of $\epsilon{=}0.28$, and a KL penalty coefficient of $\beta{=}0.001$. The learning rate is $5\times10^{-6}$ with a per-prompt batch size of 64. All DR-Rubric-8B models are initialized from Qwen3-8B-SFT and trained for 5 epochs per bootstrap step over the 1k training set on 16$\times$H800 GPUs. Bootstrap variants accumulate training across steps: BS-2 trains for 2$\times$5 epochs (2K cumulative instances), BS-3 for 3$\times$5 epochs (3K cumulative instances). Each benchmark is evaluated on $\min(400, |\mathcal{D}|)$ samples by default; Appendix~\ref{appendix:sample_size} verifies that varying the evaluation sample size (100, 400, 800) yields consistent conclusions. GPT-5 and Gemini rubric generation each required approximately 4 API calls per task.

\subsection{Rubric-Centric Evaluation Benchmarks}

\paragraph{ResearchQA.}
ResearchQA is a large-scale rubric-based benchmark designed to evaluate whether a model’s answer covers the key evaluative criteria extracted from expert-authored survey and review papers. The benchmark consists of 21.4K research questions across 75 research fields and 7 top-level domains. Each question is paired with a \emph{hybrid rubric} composed of survey-derived factual criteria, parametric knowledge-based criteria, and citation-specific checks. On average, each query is associated with approximately 7.5 rubric items, covering informational correctness, comparative analysis, depth of discussion, and explicit referencing. Evaluation is performed by computing the proportion of rubric items satisfied by the model output, yielding a fine-grained measure of structured answer completeness rather than surface-level correctness.

\subsection{Deep Research and Evidence-Grounded Benchmarks}

\paragraph{DeepResearchBench.}
DeepResearchBench evaluates deep research agents on 100 expert-designed research tasks spanning 22 domains, including science, finance, software engineering, and healthcare. Each task requires multi-step information acquisition, synthesis, and report generation. The benchmark provides two complementary evaluation frameworks. \textbf{RACE} (Reference-based Adaptive Criteria-driven Evaluation) compares generated reports against expert reference reports using adaptive, dimension-specific scoring across comprehensiveness, insight, instruction adherence, and readability. \textbf{FACT} (Factual Abundance and Citation Trustworthiness) automatically verifies whether cited sources substantiate corresponding claims, producing metrics such as citation accuracy and effective citations per task. Together, these metrics disentangle stylistic fluency from evidence-grounded reasoning.

\subsection{Domain-Constrained Expert Reasoning Benchmarks}

\paragraph{GPQA.}
GPQA (Graduate-Level Google-Proof Q\&A) is a high-difficulty multiple-choice benchmark authored by domain experts in biology, chemistry, and physics. The questions are explicitly designed to be challenging for non-experts even with internet access, while remaining answerable by specialists. The benchmark provides multiple subsets, including a high-quality expert-consensus subset (GPQA-Diamond). Evaluation is performed using accuracy and abstention rate, serving as a stress test for advanced scientific reasoning beyond memorization.

\paragraph{MMLU-Pro.}
MMLU-Pro is an enhanced version of the original MMLU benchmark, designed to improve discriminative power and robustness. It contains 12,032 multiple-choice questions across 14 domains, with each question expanded to 10 answer options to reduce guessability. Low-difficulty and noisy items are filtered using model-based screening and expert review. Evaluation focuses on accuracy under standardized prompting, with the benchmark emphasizing reasoning robustness and reduced sensitivity to prompt variations.

\paragraph{MMLU.}
MMLU (Massive Multitask Language Understanding) is a broad-coverage benchmark spanning 57 subjects across STEM, humanities, and social sciences. It contains 15,908 multiple-choice questions ranging from elementary to advanced professional difficulty. Evaluation is performed under zero-shot and few-shot settings using classification accuracy. MMLU serves as a baseline measure of general world knowledge breadth and task diversity.

\subsection{Local Agentic Search Benchmarks}

\paragraph{LocalSearchBench.}
LocalSearchBench evaluates agentic search and multi-hop reasoning in real-world local life service scenarios. The benchmark includes 300 multi-hop QA tasks constructed from a large merchant knowledge base covering dining, shopping, lodging, and lifestyle services across major cities. Tasks require reasoning over multiple constraints, such as location, price, category, and user preferences. Evaluation is conducted in a controlled environment with tool access, measuring correctness, completeness, faithfulness, fluency, and safety, as well as process-level metrics such as tool usage and interaction rounds.

\paragraph{Summary.}
Collectively, these 6 evaluation benchmarks cover rubric-level constraint satisfaction, evidence-grounded synthesis, expert-level domain reasoning, and local agentic search, enabling a holistic assessment of model behavior aligned with the structured reasoning objectives of this work.

\section{Scaling to Larger Models}
\label{appendix:scaling}

To evaluate the generality of DR-Rubric across model scales, we apply the same bootstrap pipeline to Qwen3-14B and Qwen3-30B-A3B. Table~\ref{tab:scaling_results} reports agentic task results at both scales. All DR-Rubric variants follow the identical training protocol as the 8B experiments (Section~\ref{sec:experiments}): SFT initialization followed by GRPO with bootstrap rubric rewards.

\begin{table}[h]
\caption{Agentic task results at 14B and 30B-A3B scales. \textbf{Bold} = best baseline; \underline{underline} = second best. \textcolor{green!50!black}{Green}/\textcolor{red!70!black}{red} superscripts show the gap to the best baseline.}
\label{tab:scaling_results}
\centering
\setlength{\tabcolsep}{5pt}
\renewcommand{\arraystretch}{1.05}
\footnotesize

\resizebox{0.8\textwidth}{!}{%
\begin{tabular}{l ccc}
\toprule
\rowcolor{black!4}
\textbf{Method} & \textbf{ResearchQA} & \textbf{DRBench} & \textbf{LocalSearch} \\
\midrule
\rowcolor{gray!15}
\multicolumn{4}{l}{\textbf{14B Scale}} \\
Qwen3-14B-base                     & \textbf{69.4} & \underline{38.7} & \textbf{35.7} \\
DeepSeek-R1-Distill-Qwen-14B~\citep{DBLP:journals/corr/abs-2501-12948}       & \underline{68.3} & \textbf{41.0} & \underline{35.3} \\
Ministral-3-14B-Reasoning-2512~\citep{DBLP:journals/corr/abs-2601-08584}     & 66.4 & 34.9 & \textbf{35.7} \\
WebThinker-R1-14B~\citep{webthinker}                   & 61.2 & 37.6 & 34.2 \\
\rowcolor{blue!5}
DR-Rubric-14B (BS-1)               & 73.5\up{+4.1} & 40.4\dn{-0.6} & 37.4\up{+1.7} \\
\rowcolor{blue!5}
DR-Rubric-14B (BS-2)               & 73.9\up{+4.5} & 42.9\up{+1.9} & 36.9\up{+1.2} \\
\rowcolor{blue!5}
DR-Rubric-14B (BS-3)               & 71.8\up{+2.4} & 37.7\dn{-3.3} & 38.0\up{+2.3} \\
\midrule
\rowcolor{gray!15}
\multicolumn{4}{l}{\textbf{30B-A3B Scale}} \\
Qwen3-30B-A3B                      & 67.4 & 37.5 & \textbf{37.5} \\
Tongyi-DeepResearch-30B-A3B~\citep{DBLP:journals/corr/abs-2510-24701}        & \textbf{71.7} & \textbf{41.9} & \underline{37.3} \\
WebThinker-32B-DPO~\citep{webthinker}                 & 63.1 & 37.5 & 34.8 \\
MiroThinker-1.7-mini (30B-A3B)~\citep{DBLP:journals/corr/abs-2603-15726}     & \underline{71.4} & \underline{38.7} & 36.6 \\
\rowcolor{blue!5}
DR-Rubric-30B-A3B (BS-1)          & 72.3\up{+0.6} & 40.1\dn{-1.8} & 40.0\up{+2.5} \\
\rowcolor{blue!5}
DR-Rubric-30B-A3B (BS-2)          & 73.5\up{+1.8} & 42.2\up{+0.3} & 38.3\up{+0.8} \\
\rowcolor{blue!5}
DR-Rubric-30B-A3B (BS-3)          & 72.1\up{+0.4} & 42.1\up{+0.2} & 39.0\up{+1.5} \\
\bottomrule
\end{tabular}%
}
\end{table}

At both scales, DR-Rubric consistently surpasses all baselines on ResearchQA and LocalSearchBench, confirming that the rubric-based reward framework transfers effectively across model sizes. At the 14B scale, DR-Rubric-14B (BS-2) achieves the strongest overall results (73.9 on ResearchQA, 42.9 on DRBench), outperforming scale-matched reasoning models including DeepSeek-R1-Distill-Qwen-14B and WebThinker-R1-14B. At the 30B-A3B scale, DR-Rubric-30B-A3B (BS-2) reaches 73.5 on ResearchQA and 42.2 on DRBench, surpassing Tongyi-DeepResearch-30B-A3B---a dedicated deep-research agent at the same scale. These results indicate that DR-Rubric's advantage is not specific to the 8B regime but generalizes across model capacities.

\section{Bootstrap Rubric Quality Analysis}
\label{sec:rubric_quality_analysis}

This appendix extends the aggregate rubric statistics in Table~\ref{tab:rubric_quality} (Section~\ref{subsec:rubric_quality_analysis}) with distributional details, additional specificity indicators, and semantic taxonomy that are not reported in the main text. All statistics are computed over 1,000 rubrics per step.

\subsection{Methodology}

All analyses are based on 3 $\times$ 1,000 rubrics from the bootstrap training set. Dimensions are identified via Markdown \texttt{\#\#\#\#} headers; constraints are identified as bullet-list items (``\texttt{- }'' prefix). Verifiability indicators use regex pattern matching for numerical thresholds, percentages, formulas, and citation patterns. Semantic taxonomy is assigned via keyword rules across 12 predefined categories. Cross-step similarity uses word-level Jaccard distance computed over 100 randomly sampled query pairs per step pair. Type-token ratio (TTR) is computed per rubric and averaged.

\subsection{Rubric Scale}

Table~\ref{tab:rubric_scale} reports the overall size of generated rubrics.

\begin{table}[h]
\centering
\caption{Rubric scale across bootstrap steps (mean $\pm$ std).}
\label{tab:rubric_scale}
\small
\begin{tabular}{lccc}
\toprule
\textbf{Metric} & \textbf{Step 1} & \textbf{Step 2} & \textbf{Step 3} \\
\midrule
Characters   & 2057 $\pm$ 487 & 2596 $\pm$ 753 & 2411 $\pm$ 609 \\
Words        & 286.7 $\pm$ 69.5 & 364.7 $\pm$ 108.3 & 337.3 $\pm$ 86.7 \\
Lines        & 25.5 $\pm$ 6.8 & 22.0 $\pm$ 3.9 & 21.9 $\pm$ 3.3 \\
\bottomrule
\end{tabular}
\end{table}

Step~2 produces the longest rubrics on average (2596 characters, +26.2\% over Step~1), while Step~3 partially retreats (2411 characters) but remains substantially longer than Step~1. Both later steps show higher variance ($\sigma{=}753$ and 609 vs.\ 487), reflecting a wider range of rubric complexity.

\subsection{Dimension Distribution}

\begin{figure}[h]
    \vspace{-1.2em}
    \centering
    \includegraphics[width=0.6\textwidth]{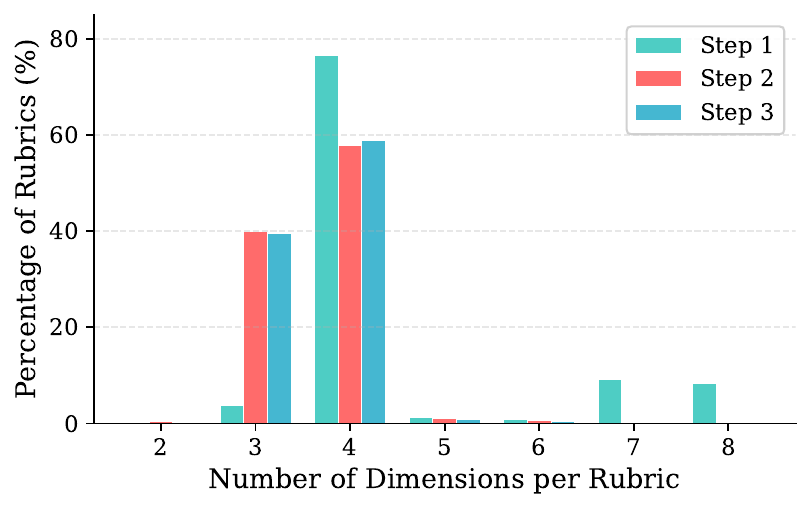}
    \vspace{-0.8em}
    \caption{Dimension count distribution across bootstrap steps. Step~1 is bimodal (peaks at 4 and 7--8 dimensions); Steps~2--3 converge to a unimodal distribution concentrated at 3--4 dimensions.}
    \vspace{-1.2em}
    \label{fig:dimension_distribution}
\end{figure}

Beyond the mean dimension counts reported in Table~\ref{tab:rubric_quality}, the distributional shape shifts qualitatively across steps. Figure~\ref{fig:dimension_distribution} visualizes this transition: Step~1 exhibits a bimodal distribution (76.6\% at 4 dimensions, 17.6\% at 7--8), which collapses into a tight unimodal peak at 3--4 dimensions in Steps~2--3 (Step~2: 57.9\% at 4, 40.0\% at 3; Step~3: 58.9\% at 4, 39.6\% at 3). Naming diversity also decreases from 3779 unique dimension names (Step~1) to $\sim$2970 (Steps~2--3), indicating convergence toward a more standardized evaluation vocabulary. The dimension range narrows from $[3,8]$ at Step~1 to $[0,6]$ at Step~3.

\subsection{Extended Specificity Indicators}

Table~\ref{tab:rubric_quality} reports two binary verifiability indicators (numerical thresholds and formulas). Table~\ref{tab:verifiability_extended} expands the analysis to five additional specificity features not covered in the main text.

\begin{table}[h]
\centering
\caption{Extended specificity indicators (\% of rubrics containing each feature). The first two rows correspond to the ``Verifiable'' and ``With formulas'' rows in Table~\ref{tab:rubric_quality}; the remaining five are additional indicators.}
\label{tab:verifiability_extended}
\small
\begin{tabular}{lccc}
\toprule
\textbf{Indicator} & \textbf{Step 1} & \textbf{Step 2} & \textbf{Step 3} \\
\midrule
\rowcolor{black!4} Numerical thresholds$^\star$ & 3.8 & 41.7 & 40.2 \\
\rowcolor{black!4} Formulas / equations$^\star$ & 14.9 & 47.3 & 46.8 \\
\midrule
Percentage data & 9.8 & 34.1 & 34.9 \\
Specific measurements & 17.9 & 52.4 & 51.1 \\
Authority references & 12.8 & 22.8 & 20.2 \\
Comparative vocabulary & 32.7 & 42.6 & 35.5 \\
Paper / author citations & 6.0 & 16.3 & 14.6 \\
\bottomrule
\end{tabular}
\vspace{0.3em}
\end{table}

The Step~1$\to$2 jump is consistent across all indicators, with specific measurements showing the broadest coverage (52.4\% at Step~2). Most indicators plateau between Steps~2 and~3, indicating that a single bootstrap iteration suffices to induce the structural shift toward executable criteria. Comparative vocabulary is the only indicator that \emph{decreases} from Step~2 to Step~3 (42.6\%$\to$35.5\%), suggesting that later steps replace comparative language with absolute thresholds.

\subsection{Dimension Semantic Taxonomy}

\begin{table}[h]
\centering
\caption{Distribution of dimension types across bootstrap steps (\% of all dimensions). Categories are assigned by keyword matching.}
\label{tab:dimension_taxonomy}
\small
\begin{tabular}{lccc}
\toprule
\textbf{Category} & \textbf{Step 1} & \textbf{Step 2} & \textbf{Step 3} \\
\midrule
Core / Primary          & 22.1 & 27.6 & 27.6 \\
Accuracy / Correctness  & 9.1 & 8.5 & 8.7 \\
Evidence / Citations     & 8.9 & 10.7 & 10.9 \\
Safety / Ethics          & 5.5 & 10.1 & 10.8 \\
Diagnosis / Clinical     & 7.8 & 5.9 & 4.8 \\
Methodology / Rigor      & 2.9 & 2.4 & 2.6 \\
Contextual / Nuance      & 3.6 & 7.3 & 8.0 \\
Risk / Limitations        & 2.4 & 2.0 & 1.5 \\
Operational / Practical   & 2.1 & 3.3 & 3.5 \\
Evaluation / Assessment   & 0.9 & 0.9 & 0.9 \\
Forbidden / Error         & 1.1 & 2.1 & 2.0 \\
Other (uncategorized)     & 31.2 & 18.2 & 17.9 \\
\bottomrule
\end{tabular}
\end{table}

Step~2 achieves higher standardization: only 18.2\% of dimension names fall outside predefined categories (vs.\ 31.2\% for Step~1), with Step~3 maintaining this gain (17.9\%). The Safety/Ethics category nearly doubles from Step~1 to Step~2 (5.5\% $\to$ 10.1\%) and continues rising at Step~3 (10.8\%), likely reflecting the influence of HealthBench medical queries in the training data. Domain-specific categories (Diagnosis/Clinical) decrease across steps (7.8\% $\to$ 5.9\% $\to$ 4.8\%) as the model generalizes toward core evaluation dimensions, while Contextual/Nuance dimensions grow substantially (3.6\% $\to$ 7.3\% $\to$ 8.0\%).

\subsection{Negation Constraint Generation}

\begin{table}[h]
\centering
\caption{Negation constraint indicators across bootstrap steps.}
\label{tab:forbidden_constraints}
\small
\begin{tabular}{lccc}
\toprule
\textbf{Metric} & \textbf{Step 1} & \textbf{Step 2} & \textbf{Step 3} \\
\midrule
Explicit ``should not'' keyword (\%) & 0.7 & 1.4 & 0.7 \\
Negation-phrased constraints (\%) & 11.0 & 34.1 & 33.6 \\
Negation constraints / rubric (mean) & 0.12 & 0.45 & 0.43 \\
\bottomrule
\end{tabular}
\end{table}

The base model (Step~1) generates relatively few rubrics containing negation constraints (11.0\%). Bootstrap training sharply increases this proportion at Step~2 (34.1\%) which plateaus at Step~3 (33.6\%), though even at Step~3 this remains below the rate observed in GPT-5-generated rubrics. This gap explains the lower proportion of negation constraints in bootstrap rubrics reported in Section~\ref{subsec:rubric_generation}.

\subsection{Cross-Step Content Independence}

The cross-step Jaccard values reported in Table~\ref{tab:rubric_quality} (0.130--0.148 for BS pairs) are elaborated here. The pairwise breakdown is: 0.130 (Step~1 vs.\ 2), 0.138 (Step~1 vs.\ 3), and 0.148 (Step~2 vs.\ 3). The slight increase across successive pairs indicates that later steps share marginally more vocabulary with each other than with Step~1, but the absolute level remains low. Combined with the near-zero constraint count correlation noted in Section~\ref{subsec:rubric_quality_analysis}, this confirms that each bootstrap step produces substantively new rubric content rather than incrementally editing previous outputs.

\section{Case Study: Detailed Evidence}
\label{appendix:case_study}

This appendix provides the full evidence supporting the case study analysis in Section~\ref{subsec:case_study}. All cases are drawn from the ResearchQA evaluation set (400 queries). The three models---Benchmark (trained with benchmark-native rubrics), GPT-5 (trained with GPT-5-generated DR rubrics), and BS-3 (trained after 3 bootstrap iterations)---share the same \texttt{plan $\to$ tavily\_search $\to$ summary} agentic pipeline. Per-item score = fraction of rubric criteria rated ``Completely.''

\subsection{Training Rubric Style Comparison}

The rubric structural properties that differentiate each model are reported in Table~\ref{tab:rubric_quality} (Section~\ref{subsec:rubric_quality_analysis}). For convenience, we highlight two additional indicators not in the main table: GPT-5 rubrics contain scoring ladders---multi-level scales (e.g., Basic/Intermediate/Advanced/Proficient/Exceptional) that map qualitative performance to discrete grade tiers---(98.3\% of rubrics) and explicit $\geq$/$>$= thresholds (93.2\%), whereas bootstrap rubrics have neither scoring ladders (0\%) nor systematic threshold syntax (BS-3: 30.5\%).

Key style differences: GPT-5 rubrics are exhaustive checklists with quantitative grading ladders ($\sim$24 constraints with 5-level scales and numerical thresholds). Benchmark rubrics are flat natural-language questions (verbose but qualitative, only 2.1\% verifiable). Bootstrap rubrics evolve: BS-1 resembles benchmark style; BS-3 develops compact dimensions with molecular-level specificity, pathway arrow notation, and explicit evidence hierarchies (verifiability 40.2\%, a $\sim$10.6$\times$ increase over BS-1).

\subsection{Case 1: Multi-Facet Query Scope (idx=291)}

\textbf{Query:} \textit{What insights do the included studies provide regarding the prevalence, clinical features, and treatment outcomes of childhood tuberculosis in Nigeria?}

\begin{table}[h]
\centering
\caption{Case~1: Scores and search queries.}
\label{tab:case1_queries}
\small
\setlength{\tabcolsep}{3pt}
\begin{tabular}{lcp{9.5cm}}
\toprule
\textbf{Model} & \textbf{Score} & \textbf{Search Query} \\
\midrule
Benchmark & 0.458 & \texttt{childhood tuberculosis Nigeria prevalence} \\
GPT-5 & 0.667 & \texttt{childhood tuberculosis Nigeria prevalence \colorbox{yellow!25}{clinical features treatment outcomes recent studies}} \\
BS-3 & \textbf{0.750} & \texttt{childhood tuberculosis Nigeria prevalence \colorbox{yellow!25}{site:.gov OR site:.org OR site:.edu}} \\
\bottomrule
\end{tabular}
\end{table}

GPT-5's multi-dimensional rubrics (6+ dimensions with independent sub-scores) incentivize covering all facets. The Benchmark's single-facet query misses treatment outcomes entirely. BS-3's domain-filtered query (\texttt{site:.gov OR site:.org}) reflects a learned emphasis on source authority from bootstrap rubrics' evidence-hierarchy constraints.

\begin{table}[h]
\centering
\caption{Case~1: Rubric criterion satisfaction.}
\label{tab:case1_rubric}
\small
\begin{tabular}{lcccc}
\toprule
\textbf{Rubric criterion} & \textbf{BM} & \textbf{GPT-5} & \textbf{BS-3} \\
\midrule
Total participants and age range across studies & N & N & N \\
Geographic/geopolitical distribution within Nigeria & N & N & N \\
Prevalence rates of childhood TB in Nigeria & \textbf{Y} & \textbf{Y} & \textbf{Y} \\
Common clinical features of childhood TB & \textbf{Y} & \textbf{Y} & \textbf{Y} \\
Treatment outcomes and success rates & N & \cellcolor{yellow!25}\textbf{Y} & \cellcolor{yellow!25}\textbf{Y} \\
Regional or demographic variations & N & N & N \\
\bottomrule
\end{tabular}
\end{table}

\subsection{Case 2: Divergent Synthesis from Identical Queries (idx=375)}

\textbf{Query:} \textit{Which clinical characteristics and treatments affect the severity and hospitalization risk of COVID-19 in people with cystic fibrosis?}

All three models produce nearly identical search queries, yet scores diverge sharply (Benchmark 0.500, GPT-5 \textbf{0.875}, BS-3 0.708). The difference lies in information extraction from retrieved content.

\begin{table}[h]
\centering
\caption{Case~2: Rubric criterion satisfaction.}
\label{tab:case2_rubric}
\small
\begin{tabular}{lcccc}
\toprule
\textbf{Rubric criterion} & \textbf{BM} & \textbf{GPT-5} & \textbf{BS-3} \\
\midrule
Specific clinical risk factors (FEV1 $<$ 40\%, post-transplant, CFRD) & N & \cellcolor{yellow!25}\textbf{Y} & \cellcolor{yellow!25}\textbf{Y} \\
Comorbidities (CF-related diabetes, pancreatic insufficiency) & N & N & N \\
Lung transplantation history as key severity factor & N & N & \cellcolor{yellow!25}\textbf{Y} \\
In-depth analysis of how clinical characteristics influence severity & N & \cellcolor{yellow!25}\textbf{Y} & N \\
Detailed explanation of specific treatments (dornase alfa, CFTR modulators) & N & \cellcolor{yellow!25}\textbf{Y} & N \\
Impact of treatment regimens on COVID-19 outcomes & \cellcolor{yellow!25}\textbf{Y} & \cellcolor{yellow!25}\textbf{Y} & N \\
\bottomrule
\end{tabular}
\end{table}

GPT-5's terse, threshold-based training constraints (4.8 words/constraint) produce a model that extracts and enumerates specific discrete facts---naming individual treatments (dornase alfa, elexacaftor-tezacaftor-ivacaftor), providing FEV1 cutoff values, and explaining mechanisms. The Benchmark model, trained on verbose yes/no constraints (26.7 words/constraint), produces topically relevant narratives but misses specific items. BS-3 hits some criteria the Benchmark misses (lung transplant history, a high-confidence risk factor) but omits others GPT-5 gets (dornase alfa), reflecting its rubric emphasis on evidence hierarchy over comprehensive enumeration.

\section{Rubric Structural Evolution: Bootstrap Self-Correction}
\label{appendix:rubric_evolution}

This appendix demonstrates how bootstrap iterations can self-correct factual errors in the rubric signal, using a clinical case with high discriminability. Table~\ref{tab:evolution_summary} summarizes how rubric structural properties evolve across sources.

\begin{table}[h]
\centering
\caption{Rubric structural evolution summary across sources (averaged over evaluation set).}
\label{tab:evolution_summary}
\small
\setlength{\tabcolsep}{3pt}
\begin{tabular}{lccccc}
\toprule
\textbf{Property} & \textbf{Bench.} & \textbf{GPT-5} & \textbf{BS-1} & \textbf{BS-2} & \textbf{BS-3} \\
\midrule
Dimensions & 0 (flat) & 6 & 7 & 4 & 4 \\
Verifiability (\%) & 2.1 & 96.4 & 3.8 & 41.7 & 40.2 \\
Negation constraints & via $-$pt & $\sim$12 & 0 & 1 & 2 \\
Scoring ladder & No & 5-level & No & No & No \\
\bottomrule
\end{tabular}
\end{table}

\subsection{Clinical Case: NVAF Stroke Prevention (Bootstrap Factual Error)}

\textbf{Query:} \textit{I wanted to have you as an AI weigh in on a stroke prevention plan for a 72-year-old patient with nonvalvular atrial fibrillation and a previous lacunar infarct. I'm a neurologist referencing the updated 2022 AHA/ASA guidelines for secondary stroke prevention \ldots\ specifically about direct oral anticoagulant recommendations.}

\smallskip\noindent\emph{Note on guideline attribution:} The query references ``updated 2022 AHA/ASA guidelines.'' The most commonly cited secondary stroke prevention guideline is the 2021 AHA/ASA Guideline for Prevention of Stroke in Patients With Stroke and TIA; AF-specific anticoagulation guidance is further addressed in the 2023 ACC/AHA/ACCP/HRS Atrial Fibrillation Guideline. The rubric texts below reproduce the models' own guideline references verbatim; factual accuracy of these attributions varies across bootstrap steps.

This case illustrates how bootstrap iterations can self-correct factual errors in the rubric signal.

\paragraph{Bootstrap Step~1 --- Factual error.}
The core dimension reads:
\begin{quote}
\small
``Mentions overlapping therapy between warfarin and DOACs as a key strategy, per the 2022 guidelines. Clarifies that bridging with low-molecular-weight heparin is no longer standard\ldots''
\end{quote}
The first constraint is factually incorrect: recommending overlapping warfarin--DOAC therapy is unsupported by and conflicts with standard anticoagulation practice for NVAF. Contemporary guidelines (2021 AHA/ASA secondary stroke prevention; 2023 ACC/AHA AF guideline) recommend direct transition from warfarin to DOACs without overlap or heparin bridging in most NVAF patients. This error reflects the noisy rubric signal inherent to Step~1, where the search-augmented generation pipeline may retrieve outdated or marginally relevant sources and the base model lacks sufficient domain grounding to filter incorrect claims.

\paragraph{Bootstrap Step~2 --- Error corrected.}
Step~2's rubric replaces the erroneous bridging claim with precise dosing criteria:
\begin{quote}
\small
``Explains the bridging protocol shift (no routine bridging with LMWH or UFH for apixaban or rivaroxaban in most NVAF patients without additional high thrombotic risk factors). Mentions standard apixaban dosing for NVAF: 5~mg twice daily for most patients; reduce to 2.5~mg twice daily only when at least two dose-reduction criteria are met---age $\geq$80 years, body weight $\leq$60~kg, or serum creatinine $\geq$1.5~mg/dL.''
\end{quote}

\paragraph{Bootstrap Step~3 --- Further refinement.}
Step~3 adds peri-procedural detail absent from Step~2:
\begin{quote}
\small
``States that for most NVAF patients undergoing temporary interruption of DOAC therapy, peri-procedural bridging is generally not recommended. Explains typical hold periods for apixaban/dabigatran/rivaroxaban/edoxaban: 48~h for creatinine clearance $>$50~mL/min; 72--120~h for reduced renal function.''
\end{quote}

\paragraph{Comparison with GPT-5.}
The GPT-5 rubric for this query contains 6 dimensions with 38 criteria and 5-level scoring ladders. Its core dimension includes 8 criteria covering DOAC preference, bridging stance, timing windows, dosing rules, and warfarin-to-DOAC transitions---each with Basic-to-Exceptional tiers. Key features absent from all bootstrap steps: (1)~explicit class-of-recommendation and level-of-evidence citations, (2)~Critical Error Gate for bridging misstatements, and (3)~trial-level citations (TIMING 2022, ELAN 2023). The Benchmark-Native rubric covers 13 weighted items with correct bridging guidance (\texttt{[+9pt, accuracy]} ``Mentions that there is no bridging protocol''; \texttt{[-8pt, accuracy]} penalizes claiming bridging is needed).

This case demonstrates two key findings: (a)~bootstrap Step~1 can introduce factual errors that propagate into training rewards, and (b)~subsequent bootstrap steps self-correct these errors as the rubric-generating model improves through GRPO training, without external supervision.

\section{Consistency Validation with Live Search API}
\label{appendix:live_api_validation}

All main experiments (Table~\ref{tab:main_results}) use a mock retrieval API backend to ensure reproducibility. To verify that the mock-based evaluation faithfully reflects real-world search conditions, we re-evaluate DR-Rubric-8B (BS-3) using the Zhipu live web search API as the tool backend. Table~\ref{tab:live_api_comparison} reports the comparison.

\begin{table}[h]
\centering
\caption{DR-Rubric-8B (BS-3) evaluated with mock API (main table) vs.\ Zhipu live web search API. All models use the same Qwen3-8B-SFT checkpoint.}
\label{tab:live_api_comparison}
\small
\setlength{\tabcolsep}{4pt}
\renewcommand{\arraystretch}{1.15}
\begin{tabular}{lcccccc}
\toprule
\textbf{Backend} & \rotatebox{70}{ResearchQA} & \rotatebox{70}{DeepResearchBench} & \rotatebox{70}{GPQA} & \rotatebox{70}{MMLU-Pro} & \rotatebox{70}{MMLU} & \rotatebox{70}{LocalSearchBench} \\
\midrule
Mock API   & 72.6 & 39.9 & 54.3 & 78.0 & 85.0 & 36.7 \\
Zhipu API  & 65.8 & 31.7 & 54.3 & 73.8 & 81.3 & 35.3 \\
\bottomrule
\end{tabular}
\end{table}

The two backends yield strongly consistent results across all 6 benchmarks. On non-retrieval-dependent benchmarks such as GPQA, MMLU, and MMLU-Pro, the scores are nearly identical (e.g., GPQA: 54.3 vs.\ 54.3), confirming that the model's reasoning capability is stable across backends. On retrieval-intensive benchmarks (ResearchQA, DeepResearchBench), the Zhipu API scores are moderately lower, which is expected: the mock API returns pre-cached, high-relevance documents, whereas a live search API introduces variability in retrieval quality and may return less targeted results. Importantly, the \emph{relative ranking} across benchmarks is preserved---benchmarks where BS-3 performs well under mock conditions remain strong under live search, and vice versa.

This consistency confirms that our mock-based experimental results are representative of real-world search-augmented inference, and that the performance patterns reported in the main paper are not artifacts of the mock retrieval backend.

\section{Training Dynamics}
\label{appendix:training_dynamics}

\begin{figure}[h]
\centering
\includegraphics[width=\textwidth]{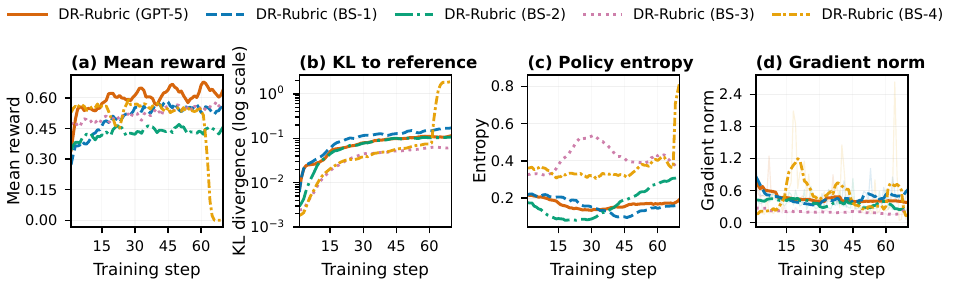}
\caption{Training dynamics over 70 GRPO steps for GPT-5 and bootstrap-rubric variants (BS-1--4).
Curves show moving averages; faint traces in gradient norm show unsmoothed values.
GPT-5-rubric training yields higher observed training reward with more gradual KL growth,
whereas bootstrap variants show stronger cumulative optimization and distinct entropy trajectories.
BS-4 exhibits training collapse: KL divergence explodes and reward degrades in late training (note broken y-axis in panel~b), demonstrating that bootstrap iteration has a natural stability limit.}
\label{fig:training_dynamics}
\end{figure}

\section{Bootstrap Stability Limit: Extended Analysis}
\label{appendix:bootstrap_stability}

\subsection{Full Per-Benchmark Results (30B-A3B, BS-1 through BS-5)}

\begin{table}[h]
\centering
\caption{30B-A3B agentic task results across bootstrap steps. BS-5 training collapsed and could not be evaluated. \textbf{Bold} = best per column.}
\label{tab:30b_full_results}
\small
\setlength{\tabcolsep}{5pt}
\renewcommand{\arraystretch}{1.15}
\begin{tabular}{lcccc}
\toprule
\textbf{Step} & \textbf{ResearchQA} & \textbf{DRBench} & \textbf{LocalSearch} & \textbf{Avg.\ Agentic} \\
\midrule
BS-1 & 72.3 & 40.1 & 40.0 & 50.8 \\
\textbf{BS-2} & \textbf{73.5} & \textbf{42.2} & 38.3 & \textbf{51.3} \\
BS-3 & 72.1 & 42.1 & \textbf{39.0} & 51.1 \\
BS-4 & 58.3 & 5.1 & 4.8 & 22.7 \\
BS-5 & \multicolumn{4}{c}{\textit{Training collapsed --- model degenerate}} \\
\bottomrule
\end{tabular}
\end{table}

Performance peaks at BS-2 (avg.\ 51.3) and remains stable at BS-3 (51.1), then collapses discontinuously at BS-4: DRBench and LocalSearch both drop by 88\% (42.1$\to$5.1 and 39.0$\to$4.8). The BS-4 model produces syntactically coherent but topically irrelevant outputs---it has overfit to reward-model surface preferences (format, keyword coverage) while losing task-solving capability. BS-5 never reaches a usable checkpoint.

\subsection{Collapse Mechanism at BS-5}

\begin{figure}[h]
\centering
\includegraphics[width=0.85\textwidth]{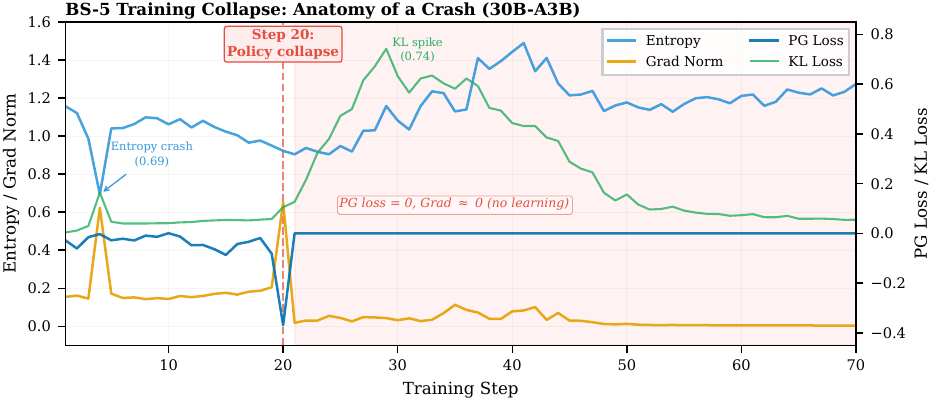}
\caption{BS-5 training collapse anatomy (30B-A3B). Policy loss spikes at step~20, gradient norm drops to near-zero, and no further learning occurs---the model is trapped at a degenerate fixed point.}
\label{fig:bootstrap_crash}
\end{figure}

BS-5's training dynamics (Figure~\ref{fig:bootstrap_crash}) exhibit a two-phase failure:

\paragraph{Phase 1: Over-optimization (steps 1--20).}
Starting from BS-4's already-polarized policy, entropy crashes to 0.69 at step~4. At step~20, policy loss spikes to $-0.37$ ($5.7\times$ BS-4's worst) with an instantaneous KL of 0.105---a single catastrophically large update that pushes the policy far from the reference distribution.

\paragraph{Phase 2: Gradient death (steps 21--70).}
Policy loss becomes exactly 0 and gradient norm collapses to ${\sim}$0.003 (from ${\sim}$0.15 pre-crash). The mechanism is specific to GRPO: all $G{=}4$ rollout samples produce near-identical outputs, yielding zero advantage variance and hence zero gradient. KL loss decays from 0.74 to 0.055 as the fixed penalty pulls the policy back toward the reference, but no task-relevant learning occurs---the model is trapped at a degenerate fixed point.

\subsection{Training Dynamics Summary}

\begin{table}[h]
\centering
\caption{Training dynamics indicators for 30B-A3B across bootstrap steps. $^\dagger$BS-5 final values reflect KL recovery from a degenerate policy, not normal convergence.}
\label{tab:30b_dynamics_summary}
\small
\setlength{\tabcolsep}{3.5pt}
\renewcommand{\arraystretch}{1.15}
\begin{tabular}{lccccc}
\toprule
\textbf{Indicator} & \textbf{BS-1} & \textbf{BS-2} & \textbf{BS-3} & \textbf{BS-4} & \textbf{BS-5}$^\dagger$ \\
\midrule
Entropy (final) & 0.88 & 1.20 & 1.28 & 1.17 & 1.27 \\
Entropy (min) & 0.23 & 0.69 & 0.62 & 0.87 & 0.69 \\
Entropy range & 0.65 & 0.51 & 0.66 & 0.30 & 0.58 \\
KL loss (final) & 0.310 & 0.094 & 0.086 & 0.066 & 0.055 \\
KL loss (max) & 0.310 & 0.094 & 0.105 & 0.082 & 0.742 \\
Grad norm (late avg) & 0.30 & 0.18 & 0.17 & 0.16 & 0.004 \\
PG loss (min) & $-0.18$ & $-0.11$ & $-0.23$ & $-0.17$ & $-0.37$ \\
PG loss (final 50 avg) & 0.003 & $-0.04$ & $-0.03$ & $-0.03$ & 0.000 \\
Polarization & 0.096 & 0.054 & 0.195 & 0.249 & 0.767 \\
Internal reward (avg) & 0.24 & 0.38 & 0.50 & 0.66 & 0.19 \\
\midrule
Agentic avg & 50.8 & \textbf{51.3} & 51.1 & 22.7 & --- \\
\bottomrule
\end{tabular}
\end{table}

Three patterns emerge from Table~\ref{tab:30b_dynamics_summary}. First, internal reward anti-correlates with external performance after BS-2: reward rises monotonically (0.24$\to$0.66) through BS-4 while agentic scores collapse, a direct instance of Goodhart's Law in iterative RLHF. Second, polarization is the most reliable early-warning signal---its $3.6\times$ jump at BS-3 (0.054$\to$0.195) precedes the BS-4 performance collapse by one full iteration, providing an actionable lead time. Third, final KL loss \emph{decreases} across stable steps (0.310$\to$0.094$\to$0.086$\to$0.066) because cumulative bootstrap initializes each step from the previous checkpoint, making KL a poor stopping criterion: it tracks parameter drift from the immediately preceding policy, not from the original base model. The BS-5 KL explosion (0.742, $8.9\times$ BS-4's maximum of 0.082) confirms that the step-20 update was catastrophically out-of-distribution.

\subsection{Cross-Scale Comparison: 14B vs.\ 30B-A3B}

\begin{table}[h]
\centering
\caption{Bootstrap over-optimization across model scales. Both degrade after BS-2, but through different mechanisms.}
\label{tab:cross_scale_collapse}
\small
\setlength{\tabcolsep}{3.5pt}
\renewcommand{\arraystretch}{1.15}
\begin{tabular}{lcc}
\toprule
\textbf{Dimension} & \textbf{14B} & \textbf{30B-A3B} \\
\midrule
Optimal step & BS-2 & BS-2 \\
First degradation step & BS-3 & BS-3 \\
Polarization at optimal & 0.026 & 0.054 \\
Polarization jump ratio (opt$\to$next) & $2.4\times$ & $3.6\times$ \\
Failure mode & Capacity saturation & Reward hacking \\
Internal reward trend & Decreasing & Increasing \\
Token corruption & None & Severe (Korean injection, random tokens) \\
Grad norm stability & Stable throughout & Collapses at BS-5 \\
Stopping criterion fires at & BS-3 (correct) & BS-3 (correct) \\
\bottomrule
\end{tabular}
\end{table}

The 14B model degrades because it exhausts representational capacity (reward \emph{decreases}, violations increase); the 30B model degrades because it exploits the reward model (reward \emph{increases}, task quality decreases). Despite these opposite causal mechanisms, polarization detects both at BS-3 because it captures a shared distributional symptom---extreme-score concentration---rather than a mechanism-specific signal.

\subsection{Practical Stopping Guideline}

\begin{enumerate}
\item \textbf{Primary criterion:} After bootstrap step $n$, compute polarization $P_n$ (fraction of samples scoring 0 or $\geq$0.99). Stop if $P_n > \max(0.15,\; 2 \times P_{n-1})$.
\item \textbf{Model selection:} $\arg\min_k P_k$ (lowest polarization).
\item \textbf{Auxiliary:} If $\min(\text{entropy}) < 0.70$ within a step, the rubrics are too narrow---consider reducing constraint specificity.
\item \textbf{Hard bound:} Do not exceed 3 bootstrap iterations without external rubric grounding (e.g., a GPT-5 ``reset'' step).
\end{enumerate}

Retrospective validation on both scales: this rule correctly selects BS-2 and terminates at BS-3, avoiding BS-4/BS-5 and saving 40--60\% of total bootstrap compute.

\section{Fresh-Start Bootstrap: Disentangling Rubric Quality from Policy Drift}
\label{appendix:fresh_start_bootstrap}

The cumulative bootstrap pipeline conflates two factors: improved rubric quality and accumulated policy optimization. To isolate the rubric contribution, we introduce a \emph{fresh-start} variant that re-initializes from Qwen3-8B-SFT at each step:
\begin{equation}
\pi_{\theta_{t+1}}^{\text{fresh}} = \text{GRPO}(\pi_{\text{base}},\ \mathcal{R}_{p,t}), \quad \text{vs.} \quad \pi_{\theta_{t+1}} = \text{GRPO}(\pi_{\theta_t},\ \mathcal{R}_{p,t}).
\end{equation}
All other hyperparameters remain identical; only the policy initialization differs.

\begin{table}[h]
\centering
\caption{Fresh-start vs.\ cumulative bootstrap. Both variants use identical rubrics at each step; only the policy initialization differs. All models use Qwen3-8B-SFT as base.}
\label{tab:fresh_start_bootstrap}
\small
\setlength{\tabcolsep}{4pt}
\renewcommand{\arraystretch}{1.15}
\begin{tabular}{l >{\columncolor{blue!5}}c >{\columncolor{blue!5}}c >{\columncolor{blue!5}}c >{\columncolor{orange!7}}c >{\columncolor{orange!7}}c >{\columncolor{orange!7}}c}
\toprule
\rowcolor{black!4}
& \multicolumn{3}{c}{\cellcolor{blue!5}\textit{Agentic Task}} & \multicolumn{3}{c}{\cellcolor{orange!7}\textit{Expert Reasoning}} \\
\cmidrule(lr){2-4} \cmidrule(lr){5-7}
\rowcolor{black!4}
\textbf{Method} & ResearchQA & DRBench & LocalSearch & GPQA & MMLU-Pro & MMLU \\
\midrule
Cumulative BS-2  & 72.4 & 38.3 & 36.2 & 56.5 & 74.3 & 83.5 \\
Fresh-start BS-2 & 72.8 & 39.6 & 35.5 & 53.5 & 76.8 & 82.0 \\
\midrule
Cumulative BS-3  & 72.4 & 39.5 & 36.4 & 55.8 & 78.0 & 85.3 \\
Fresh-start BS-3 & 72.6 & 39.9 & 36.7 & 54.3 & 78.0 & 85.0 \\
\bottomrule
\end{tabular}
\end{table}

Fresh-start BS-3 closely matches cumulative BS-3 across all six benchmarks despite never benefiting from prior RL training, demonstrating that bootstrap gains are driven by rubric quality evolution rather than accumulated policy optimization. Both variants exhibit the same BS-2 dip, confirming that the non-monotonic trajectory originates from rubric$^{(2)}$'s structural properties (the verifiability jump from 3.8\% to 41.7\%) rather than cumulative optimization. This also rules out circular optimization concerns: rubrics generated by $\pi_{\theta_t}$ are equally effective on a fresh base model, indicating they encode transferable task knowledge rather than policy-specific reward artifacts.

\section{Evaluation Sample Size Sensitivity}
\label{appendix:sample_size}

Our default evaluation protocol uses $\min(400, |\mathcal{D}|)$ samples per benchmark. To verify that this budget does not bias conclusions, we re-evaluate DR-Rubric-8B (BS-3) with 100 and 800 evaluation samples on the same 6 benchmarks used in the main results. Table~\ref{tab:sample_size} reports the comparison.

\begin{table}[h]
\centering
\caption{Evaluation sample size sensitivity for DR-Rubric-8B (BS-3). All runs use the same Qwen3-8B-SFT checkpoint; only the number of evaluation samples differs.}
\label{tab:sample_size}
\small
\setlength{\tabcolsep}{4pt}
\renewcommand{\arraystretch}{1.15}
\begin{tabular}{l >{\columncolor{blue!5}}c >{\columncolor{blue!5}}c >{\columncolor{blue!5}}c >{\columncolor{orange!7}}c >{\columncolor{orange!7}}c >{\columncolor{orange!7}}c}
\toprule
\rowcolor{black!4}
& \multicolumn{3}{c}{\cellcolor{blue!5}\textit{Agentic Task}} & \multicolumn{3}{c}{\cellcolor{orange!7}\textit{Expert Reasoning}} \\
\cmidrule(lr){2-4} \cmidrule(lr){5-7}
\rowcolor{black!4}
\textbf{Sample Size} & ResearchQA & DRBench & LocalSearch & GPQA & MMLU-Pro & MMLU \\
\midrule
100  & 70.4 & 39.3 & 36.2 & 51.0 & 75.0 & 80.0 \\
400 (default) & 72.6 & 39.9 & 36.7 & 54.3 & 78.0 & 85.0 \\
800  & 71.5 & 39.8 & 36.7 & 52.7 & 76.9 & 84.3 \\
\bottomrule
\end{tabular}
\end{table}

The three sample sizes yield consistent performance patterns: agentic task benchmarks (ResearchQA, DRBench, LocalSearchBench) remain stable across all budgets, while expert-reasoning benchmarks (GPQA, MMLU-Pro, MMLU) show moderate variance due to sampling noise. Per-benchmark scores remain within narrow ranges across sample sizes---confirming that our evaluation budget is sufficient for reliable conclusions and that the main findings are not artifacts of a particular sample size.

\section{Rubric Granularity Ablation: Full Results}
\label{appendix:rubric_granularity}

Table~\ref{tab:rubric_granularity_full} reports the full per-benchmark breakdown for the rubric granularity ablation summarized in Table~\ref{tab:rubric_granularity}. All variants use DR-Rubric-8B (BS-1) with identical training setup; only the rubric generation parameters ($n_{\text{dim}}$, $n_{\text{cons}}$) differ.

\begin{table}[h]
\centering
\caption{Rubric granularity ablation: full per-benchmark results. Dim./Cons.\ = dimension range / constraint range used during rubric generation. \textbf{Bold} = best per column.}
\label{tab:rubric_granularity_full}
\small
\setlength{\tabcolsep}{4pt}
\renewcommand{\arraystretch}{1.15}
\begin{tabular}{l >{\columncolor{blue!5}}c >{\columncolor{blue!5}}c >{\columncolor{blue!5}}c >{\columncolor{orange!7}}c >{\columncolor{orange!7}}c >{\columncolor{orange!7}}c}
\toprule
\rowcolor{black!4}
& \multicolumn{3}{c}{\cellcolor{blue!5}\textit{Agentic Search}} & \multicolumn{3}{c}{\cellcolor{orange!7}\textit{Expert Reasoning}} \\
\cmidrule(lr){2-4} \cmidrule(lr){5-7}
\rowcolor{black!4}
\textbf{Rubric Size} (Dim./Cons.) & ResearchQA & DRBench & LocalSearch & GPQA & MMLU-Pro & MMLU \\
\midrule
Compact (2--4 / 5--8)    & 70.2 & 39.2 & 36.7 & \textbf{57.0} & \textbf{77.5} & \textbf{84.0} \\
Medium (7--10 / 15--25)  & 62.2 & 37.3 & 36.7 & 52.0 & 74.0 & 80.8 \\
Large (10--14 / 25--30)  & 70.3 & 41.7 & 36.3 & 55.5 & 71.3 & 83.0 \\
X-Large (15--20 / 40--50) & \textbf{71.9} & \textbf{41.6} & \textbf{38.3} & 54.0 & 73.3 & 79.8 \\
\bottomrule
\end{tabular}
\end{table}

Agentic search benchmarks (ResearchQA, DRBench, LocalSearchBench) improve consistently with rubric breadth: X-Large rubrics achieve the highest scores on ResearchQA (71.9) and LocalSearchBench (38.3), and are competitive on DRBench (41.6). In contrast, expert reasoning benchmarks (GPQA, MMLU-Pro, MMLU) uniformly favor compact rubrics: the Compact setting achieves the best scores on all three (57.0, 77.5, 84.0). The per-category trends are clearly divergent---reinforcing the finding from the main text that rubric structure mediates a breadth--depth trade-off across task families.

%%%%%%%%%%%%%%%%%%%%%%%%%%%%%%%%%%%%%%%%%%%%%%%%%%%%%%%%%%%%

% \input{contents/paper_checklist}

\end{document}